\RequirePackage{fix-cm}
\documentclass[smallcondensed]{svjour3}     
\smartqed  
\usepackage{amsmath}
\usepackage{amssymb}
\usepackage{array}
\usepackage{color}
\usepackage{cases}
\usepackage{hyperref}
\usepackage{cite}
\usepackage{algorithmic}
\usepackage{pifont}
\usepackage{algorithm}
\usepackage{graphicx, color}
\usepackage{epstopdf}
\usepackage{booktabs}
\usepackage{diagbox}
\usepackage{multirow, multicol}
\usepackage{arydshln}
\usepackage{subfigure}
\usepackage[misc]{ifsym}
\graphicspath{{figs/}}

\newcommand \red [1]{\textcolor{red}{#1}} 
\newcolumntype{L}[1]{>{\raggedleft\let\newline\\\arraybackslash\hspace{0pt}}m{#1}}
\newcolumntype{R}[1]{>{\raggedleft\let\newline\\\arraybackslash\hspace{0pt}}m{#1}}
\usepackage{geometry}
\usepackage{setspace}
\usepackage[inline, shortlabels]{enumitem}
\setlist[enumerate]{leftmargin=.5in}
\setlist[itemize]{leftmargin=.5in}

\begin{document}
\title{
Enhancing Electrical Impedance Tomography reconstruction using Learned Half-Quadratic Splitting Networks with Anderson Acceleration}
\author{Guixian Xu \textsuperscript{1} \and Huihui Wang \textsuperscript{1} \and {Qingping Zhou \textsuperscript{1, \Letter}}}
\institute{%
\begin{itemize}[leftmargin=*]
  \item[\textsuperscript{\Letter}] {Qingping Zhou} \\
        \email{qpzhou@csu.edu.cn} 
        \\
  \item[] {Guixian Xu} \\
        \email{xuguixian@csu.edu.cn}
        \\
  \item[] {Huihui Wang} \\
        \email{huihuiwang@csu.edu.cn}
  \at
  \item[\textsuperscript{1}] School of Mathematics and Statistics, HNP-LAMA, Central South University, Changsha, China
\end{itemize}
}
\titlerunning{Enhancing EIT reconstruction using AA-HQSNet}

\date{}
\maketitle
\begin{abstract}
Electrical Impedance Tomography (EIT) is widely applied in medical diagnosis, industrial inspection, and environmental monitoring. 
Combining the physical principles of the imaging system with the advantages of data-driven deep learning networks, physics-embedded deep unrolling networks have recently emerged as a promising solution in computational imaging. 
However, the inherent nonlinear and ill-posed properties of EIT image reconstruction still present challenges to existing methods in terms of accuracy and stability.
To tackle this challenge, we propose the learned half-quadratic splitting (HQSNet) algorithm for incorporating physics into learning-based EIT imaging. We then apply Anderson acceleration (AA) to the HQSNet algorithm, denoted as AA-HQSNet, which can be interpreted as AA applied to the Gauss-Newton step and the learned proximal gradient descent step of the HQSNet, respectively. 
AA is a widely-used technique for accelerating the convergence of fixed-point iterative algorithms and has gained significant interest in numerical optimization and machine learning. However, the technique has received little attention in the inverse problems community thus far. 
Employing AA enhances the convergence rate compared to the standard HQSNet
while simultaneously avoiding artifacts in the reconstructions.
Lastly, we conduct rigorous numerical and visual experiments to show that the AA module strengthens the HQSNet, leading to robust, accurate, and considerably superior reconstructions compared to state-of-the-art methods.
Our Anderson acceleration scheme to enhance HQSNet is generic and can be applied to improve the performance of various physics-embedded deep learning methods.
\end{abstract}

\clearpage
\section{Introduction}\label{sec:intro}
Electrical impedance tomography (EIT) is a radiation-free imaging technique that enables repetitive and non-invasive measurement of regional changes within an object, making it useful for various applications~\cite{adler2021electrical}, such as medical diagnosis, industrial inspection, and environmental monitoring. 
EIT image reconstruction is usually cast as an inverse problem as one wants to determine the input to a system from its output. The problem has attracted considerable research interest, and a number of methods have been developed to recover the image, e.g., \cite{hamilton2018deep,seo2019learning,adler2021electrical,9806671,colibazzi2022learning}, just to name a few.

In the ElT image reconstruction problem, the number of unknowns can exceed the number of observations, and the measurement and recording process is inevitably corrupted by noise. Moreover, the inputs (electrical currents and potentials), outputs (electrical conductivity), and system parameters exhibit complicated nonlinear interrelations. These factors render the EIT reconstruction problem ill-posed in the sense that the solution may not exist, be not unique, or not vary smoothly on the data due to small noise variations. Consequently, the EIT reconstruction problem cannot be adequately addressed using standard dimension reduction or regularized regression techniques.
To address the challenges mentioned above, numerous methods have been developed, with prototypical examples including linear back projection~\cite{gamio2003interpretation}, Landweber iterative algorithm~\cite{liu2012landweber}, non-iterative reconstruction method~\cite{ferreira2017new}, D-Bar reconstruction~\cite{isaacson2004reconstructions}, regularization techniques~\cite{jin2012analysis,gonzalez2017isotropic,shi2019reduction}, end-to-end deep learning~\cite{michalikova2014image,cheng2022r}, post-processing deep learning methods~\cite{hamilton2018deep,hamilton2019beltrami,wei2019dominant,9806671}, and physics-embedded deep unrolling networks~\cite{colibazzi2022learning}.  
Combining the strengths of the physical principle of the imaging process and data-driven deep learning solutions, physics-embedded deep unrolling networks have gained popularity in many areas of inverse problems due to their state-of-the-art performance. Although they have been well studied for the linear inverse problems~\cite{zhang2020deep,xin2022learned,tang2022accelerating,sahel2022deep}, the learning methods for the EIT imaging inverse problem vary widely ~\cite{zhang2020supervised,colibazzi2022learning,guo2023physics} due to its intrinsic nonlinear and ill-posed properties. 
Further discussion on related work will be presented in Section~\ref{sec:relatedwork}.

We aim to develop a comprehensive framework for performing the learned unrolling half-quadratic splitting method with Anderson acceleration (AA-HQSNet) for the EIT reconstruction problem.
Anderson acceleration, also known as Anderson mixing or Pulay mixing, has been rediscovered several times in different communities over the last 50 years. It improves the convergence of fixed-point or optimization problems by utilizing a linear combination of previous estimates. We refer interested readers to~\cite{anderson2019comments} for a thorough review of this topic. 
While the notion of Anderson acceleration (AA) has gained significant interest in numerical optimization~\cite{zhang2020globally,bollapragada2022nonlinear} and machine learning~\cite{geist2018anderson,shi2019regularized}, but so far it has received little attention in the inverse problems community.
Some initial research in this intersection has been pioneered by 
~\cite{mai2020anderson,pasini2021stable,pasini2022anderson}.

We summarize the key ideas of our proposed framework as the following. 
First, we introduce the learned half-quadratic splitting network (HQSNet) for the EIT problem. HQSNet divides the problem into two subproblems: the Gauss-Newton and the learned proximal gradient descent steps. 
Using the deep unrolling method, we avoid the need for manual algorithm parameter selection, such as $\lambda$ in~\eqref{0}, and can achieve higher generalization compared to the classical iterative optimization algorithms or deep learning method.
Second, we consider the numerical implementation of HQSNet. 
For this high-dimensional EIT problem,  the computation of the Jacobian matrix is computationally expensive. However, HQSNet only uses the current estimate and its gradient information, ignoring a large number of historical estimates and their Jacobian information, resulting in a waste of valuable information. 
To address this issue, in~\cite{ye2020nesterov} Zhang and Ye introduce  Nesterov's acceleration technique to improve the efficiency of past information usage. However, our work adopts an alternative approach based on the Anderson acceleration (AA) technique.
We note that similar methods are proposed in~\cite{mai2020anderson,pollock2020benchmarking}. The main difference between our approach and~\cite{pollock2020benchmarking} is that we use Gauss-Newton optimization, where the Hessian matrix is approximated using the Jacobian matrix, while the authors of~\cite{pollock2020benchmarking} assume that the Hessian matrix is already known when using the standard Newton method. Therefore, approximation of the Hessian is not an issue in their approach. 
For the work by Mai and Johansson~\cite{mai2020anderson}, we use a parameterized neural network to replace the explicit proximal gradient operator in~\cite{mai2020anderson} that is preferred for theoretical convergence in~\cite{mai2020anderson}.
Finally, we provide a large number of numerical experiments and algorithm investigations to verify the effectiveness and efficacy of our proposed approach.
The proposed approach has several appealing features. 
Firstly, it automatically and simultaneously addresses multiple challenges in the EIT problem, such as allowing for the automatic learning of hyperparameters and learning a data-driven regularizer. 
Secondly, it preserves the physical principle of the imaging system, which enhances the learning process and improves the generalization ability. 
Finally, it exhibits high accuracy and stability to noise, making it suitable for handling noisy data in real-world applications.

The paper is structured as follows. Section~\ref{sec:eit} provides an overview of the EIT image reconstruction problem, along with a review of related literature. 
In Section~\ref{sec:method}, we introduce AA-HQSNet, an algorithm that combines Anderson acceleration with learned half-quadratic splitting to solve the EIT inverse problem. Section~\ref{sec:experiment} presents the main experimental results, and we further investigate the proposed algorithm in Section~\ref{sec:algo-analysis}. Finally, Section~\ref{sec:conclusion} offers concluding remarks.

\section{Electrical Impedance Tomography}\label{sec:eit}
\subsection{EIT forward and inverse problem}
In this paper, we consider a bounded domain $\Omega \subseteq \mathbb{R}^{m}, m=2$ with a $C^1$ boundary that contains certain conducting materials. The electrical conductivity of these materials is defined by a positive function $\sigma(x) \in L^{\infty}(\Omega)$.
The homogeneous background material has a conductivity of $\sigma_0$, and the inhomogeneous inclusions indicated by $D$ are determined by the support of $\sigma-\sigma_0$.
We assume that $L$ different electrical currents are injected into the boundary of $\partial \Omega$ and the resulting electrical potential should satisfy the following $L$ physical equations with the same coefficient but different boundary conditions:
\begin{equation}\label{eq:eit_pde}
\left\{\begin{array}{ll}\nabla \cdot(\sigma \nabla u)=0 & \text { in } \Omega, \\ 
u+z_{l} \sigma \frac{\partial u}{\partial n}=v_{l} & \text { on } E_{l}, l=1, \ldots, L, \\ 
\int_{E_{l}} \sigma \frac{\partial u}{\partial n} \mathrm{~d} s=I_{l} & \text { on } \Gamma, \\ 
\sigma \frac{\partial u}{\partial n}=0 & \text { on } \tilde{\Gamma},\end{array}\right.
\end{equation}
where \(\Gamma(\tilde{\Gamma})\) is the boundary of \(\partial \Omega\) with (without) electrodes, \(v_{l}\) is the voltage measured by \(l\)-th electrode \(E_{l}\) when the currents \(I_{l}\) are applied, \(z_{l}\) is the contact impedance, $u$ is the electric potential in the interior of the object $\Omega$, and $n$ is the normal to the object surface. \par
\begin{figure}[htp]
\centering
\includegraphics[width=0.6\textwidth]{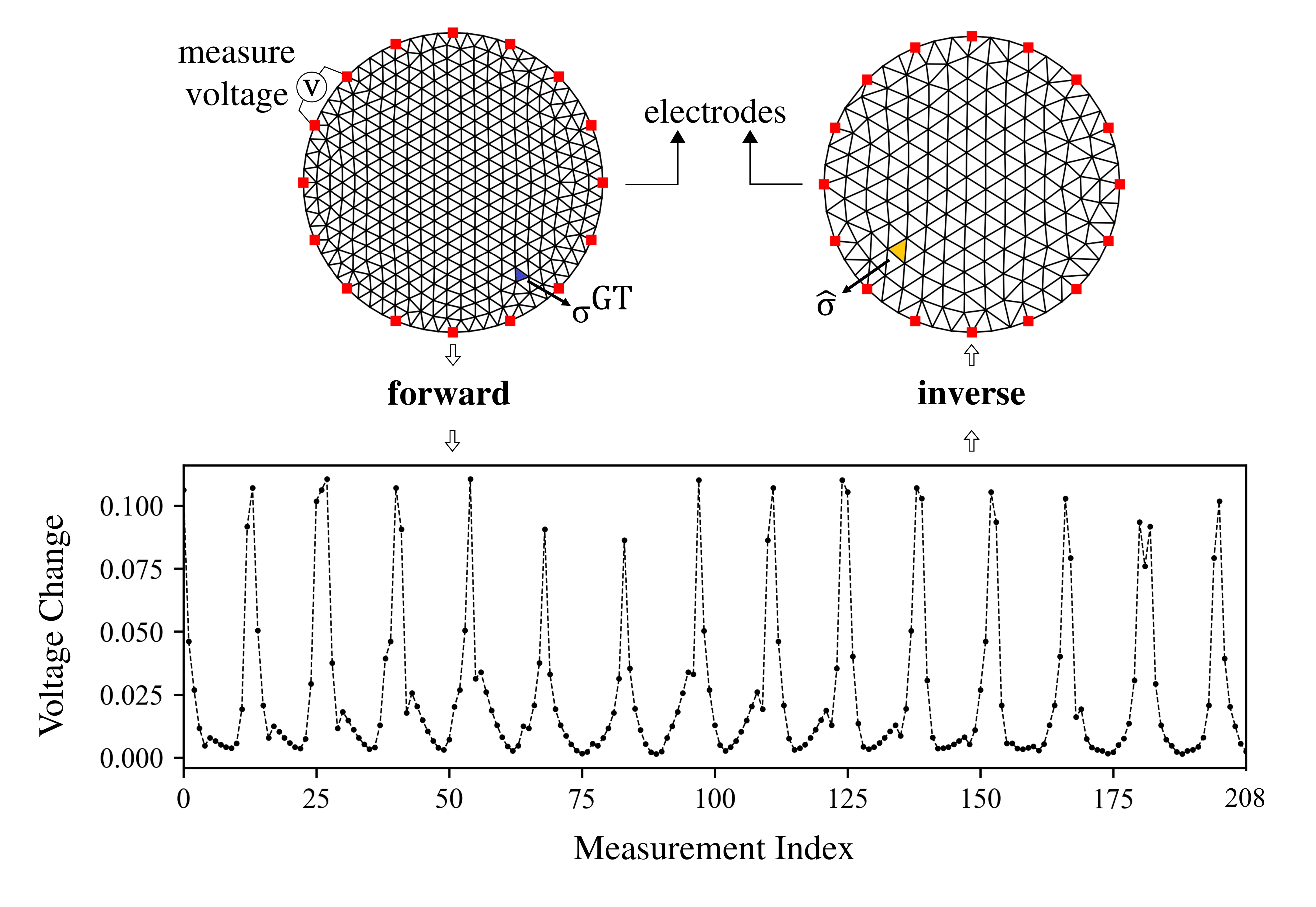}
\caption{EIT forward and inverse scenarios. After the surface of the object is represented as a finite element mesh, the forward process transforms the mesh data into a measured voltage value; the inverse process then back-performs the reconstruction of ground truth conductivity value from this measured voltage}
\label{fig:eit}
\end{figure}

The EIT forward problem involves to calculating the estimates for the electric potential in $\Omega$ assuming $\sigma$ is known. Operating between the Hilbert spaces $X$ and $Y$, the \textit{Forward Operator $\tilde{F}$} maps the conductivity $\sigma$ to the solution of the forward problem:
\begin{equation*}
\begin{aligned}
    \tilde{F}: \Theta \subset X &\to Y\\
    \sigma &\mapsto (u, V)
\end{aligned}
\end{equation*}
where $\Theta = \left\{\sigma \in L^{\infty}(\Omega) | \sigma \nabla u = 0\right\}$ denotes the domain of definition of $\tilde{F}$. The EIT forward problem can be solved by representing the surface as a finite element mesh, and the object domain $\Omega$ is therefore divided into a discrete set of $n_{T}$ subdomains $\left\{\tau_{j}\right\}_{j=1}^{n_{T}}$ (see Fig.~\ref{fig:eit}), and $\sigma$ is constant over each subdomain $\tau_j$. 
Given a finite element model of an EIT medium, we calculate the vector of voltages, $v$, for each degree of freedom in the model. 
To obtain a vector of $n_{M}$ measurements, we can apply a given stimulation pattern by injecting current through a pair of electrodes and measuring the corresponding voltage $v$ induced on another pair of electrodes. $F: \mathbb{R}^{n_{T}} \rightarrow \mathbb{R}^{n_{M}}$
represents the discrete version of the \textit{Forward Operator}, and $\tilde{F}$ is Fr$\acute{e}$chet differentiable, as for the $F^{'}$, denoted by $J$, is the Jacobian of $F$; each element of $J \in \mathbb{R}^{n_{M} \times n_{T}}$ is defined as
\begin{equation}\label{joca}
\left\{ J(u_{d}, u_{m}) \right\}_{i,j} = \frac{\partial v_{i}}{\partial \sigma_{j}} =\int_{\tau_{j}} \nabla u_{d} \cdot \nabla u_{m} d\Omega,
\end{equation}
where the row indicator $i$ refers to the $i$-th measurement, which is obtained from the $d$-th emitting electrode and the $m$-th measuring electrode, and $j$ corresponds to the subdomain $\tau_{j}$.
Considering the presence of additive noise in measured data, we define the noisy nonlinear observation model as follows:
\begin{equation}\label{eq:eit_nonlinear}
v=F(\sigma)+\eta \text {, }
\end{equation}
where $v \in \mathbb{R}^{n_{M}}$ denotes the vector of all the measured electrode potentials, whose dimension $n_{M}$ depends on the choice of the measurement protocol. Moreover, $\eta \in \mathbb{R}^{n_{M}}$ represents a zero-mean Gaussian-distributed measurement noise vector.

Mathematically, the EIT inverse problem involves recovering the conductivity $\sigma$ within the domain $\Omega$ based on the measurements $v$ and the nonlinear mapping $F(\sigma)$. The purpose of the so-called absolute imaging problem is to estimate the conductivity $\sigma$ by solving the following non-linear least squares problem:
\begin{equation}\label{6}
\hat{\sigma}\in\mathop{\arg\min}\limits_{\sigma} \frac{1}{2} \Vert v - F(\sigma) \Vert_2^{2} + \lambda R(\sigma),
\end{equation}
where $\frac{1}{2} \Vert v - F(\sigma) \Vert_2^{2}$ is the data fidelity term, $R(\sigma)$ is the regularization term, and $\lambda$ is a trade-off parameter. The EIT forward and inverse problems are widely recognized as challenging. The forward problem is a map from an $n_T$-dimensional function to a $2(n_T-1)$-dimensional function. The EIT inverse problem is widely recognized as challenging due to its severe ill-posedness.

\subsection{Related Work}\label{sec:relatedwork}
As discussed in Section~\ref{sec:intro}, various approaches have been proposed to solve the EIT image reconstruction problem. We will provide a brief overview of the two most relevant techniques to this work: regularization techniques and deep learning-based methods.
\paragraph{\textbf{Regularization techniques}} 
Regularization techniques aim to stabilize the ill-posed property of the EIT problem by adding constraints to the solution space. These methods formulate a data consistency and a regularizer-composed variational problem and solve it with iterative algorithms, such as the alternating direction method of multipliers~\cite{boyd2011distributed}, and the half-quadratic splitting (HQS)~\cite{geman1995nonlinear}. In the context of EIT reconstruction, a variety of regularizers have been employed, such as Tikhonov~\cite{jin2012analysis}, total variation~\cite{gonzalez2017isotropic} and its variations~\cite{shi2019reduction, huska2020spatially}, as well as $l_1$~\cite{dai2008electrical} regularization. 
Although these regularized techniques can provide stable approximations of the true solutions, they have a number of limitations, including slow convergence, the requirement of parameter tuning, and mathematical inflexibility.
\paragraph{\textbf{Deep learning methods}}
Another influential class of methods for image reconstruction is deep learning approaches, which completely differ from regularized data-fitting methods that use a single data fidelity with regularization. Deep learning-based reconstruction approaches trained on paired data are known to have a highly expressive representation that captures anatomical geometry as well as small anomalies~\cite{seo2019learning}. We group the existing deep learning methods for EIT image reconstruction  into three categories: post-processing learned approaches, purely deep learning-based methods, and physics-embedded deep unrolling networks.
Firstly, post-processing learned approaches utilize a trained network to eliminate artifacts of a preliminary conductivity image obtained from model-based algorithms~\cite{hamilton2018deep,hamilton2019beltrami,wei2019dominant,9806671}.
In~\cite{hamilton2018deep,hamilton2019beltrami}, they proposed combining the D-bar algorithm with U-Net to improve the sharpness and clarity of reconstructed EIT images. 
Wang et al.~\cite{9806671} use a multi-channel post-process convolution neural network to modify the initial medium distribution obtained by Newton-Raphson iteration methods. However, the effectiveness of these methods is closely tied to the success of the first reconstruction phase since deep learning is not utilized to incorporate data-driven knowledge into the regularization process.
Secondly, purely learning-based methods use parameterized neural networks to learn the linear or nonlinear mapping from the observation to the target image, see e.g.,~\cite{michalikova2014image,cheng2022r,9806671}. Data-based learning is favorable when the imaging mechanism is unclear or has significant errors compared to the true physical process. Despite its approximation to the true physical problem, the forward operator still contains considerable information, rendering the disregard of its internal mechanism suboptimal. Consequently, there is a current research emphasis on integrating mechanisms and data in image inverse problems.
Thirdly, physics-embedded deep unrolling networks explicitly unroll iterative optimization algorithms into learnable deep architectures. 
This approach began with the work of ~\cite{gregor2010learning}, where they applied the unrolling framework to the iterative shrinkage-thresholding algorithm. 
Embedding physics models into neural networks has demonstrated empirical success in implementing a wide range of classical iterative optimization algorithms, such as the unrolled alternating directions method of multipliers ~\cite{boyd2011distributed}, hybrid gradient primal-dual method~\cite{chambolle2011first}, Gauss-Newton~\cite{colibazzi2022learning} and half-quadratic splitting (HQS)~\cite{geman1995nonlinear}. 

\section{Methodology}\label{sec:method}
In this section, we discuss the details of our Anderson Acceleration embedded Learned Half-Quadratic Splitting (AA-HQSNet) algorithm. 
First, we introduce the learned half-quadratic splitting (HQSNet) approach, which is guided by the physical model of EIT imaging. 
Second, we present an overview of Anderson acceleration (AA) for solving optimization problems. 
Finally, we describe the integration of AA with HQSNet to develop the proposed AA-HQSNet method.

\subsection{HQSNet}\label{sec:hqs-net}
The learned half-quadratic splitting (HQSNet) algorithm has shown significant empirical success in solving optimization problems involving total variation regularization. We hereby briefly describe the corresponding HQSNet method to our problem following~\cite{xin2022learned}. 
Introducing an auxiliary variable $z$, we rewrite the optimization problem~\eqref{6} as
\begin{equation}\label{0}
\min_{\sigma,z} \frac{1}{2} \Vert v - F(\sigma) \Vert^{2} + \lambda R(z) \ \ \ s.t.  \ \ z=\sigma.
\end{equation}
We can reformulate~\eqref{0} as
\begin{equation}\label{7}
\min_{\sigma, z} \frac{1}{2{\tiny}} \Vert v - F(\sigma) \Vert^{2} + \lambda R(z) + \frac{\mu}{2} \Vert z - \sigma \Vert^{2},
\end{equation}
where $\mu>0$ is a constant known as the penalty parameter. The resulting  problem is then solved with the HQS:
\begin{subequations}
\begin{numcases}{}
\sigma_{k}=\mathop{\arg\min}\limits_{\sigma} \ \frac{1}{2} \Vert v - F(\sigma) \Vert^{2} +  \frac{\mu}{2} \Vert z_{k-1} - \sigma \Vert^{2},\label{8}  \\
z_{k}=\mathop{\arg\min}\limits_{z} \ \frac{\mu}{2}\Vert \sigma_{k}-z \Vert^{2} + \lambda R(z). \label{9} 
\end{numcases}
\end{subequations}
The algorithm consists of a $\sigma$-minimization~\eqref{8} step and a $z$-minimization step~\eqref{9}. For the $\sigma$ sub-problem, the authors~\cite{xin2022learned} give a closed-form solution under the condition that the forward operator $F(\cdot)$ is a linear operator in the MRI reconstruction problem. Here, we provide an alternative approach using the Gauss-Newton algorithm. For the $z$ sub-problem, proximal gradient descent is used to solve the $z$-minimization step. Details are discussed in the next section.

We replace the proximal operators in the $z$ sub-problem with independent parameterized neural networks $\Phi_{\theta}$, where $\theta$ represents the learned parameters, resulting in the learned HQS method (HQSNet). The complete HQSNet is sketched in Algorithm~\ref{alg:hqs}. Given a predetermined number of iterations $K$, HQSNet can be interpreted as a neural network that generates a sequence of approximated solutions $\sigma_{k} (k=1, \cdots, K)$ initialized by $\sigma_{0}$:
\begin{equation*}
\begin{cases}
    \sigma_{k}=\mathcal{G}(\sigma_{k-1};J,v,z_{k-1},\mu)  \\ 
    z_{k}=\Phi_{\theta}(\verb|c|(z_{k-1}, \frac{\partial \mathcal{L}}{\partial z_{k-1}})),  
\end{cases}
\end{equation*}
where $\mathcal{L} = \frac{\mu}{2}\Vert \sigma_{k}-z \Vert^{2}$.

\begin{algorithm}[htb!]
\begin{spacing}{1.2}
\caption{The learned half-quadratic splitting method (HQSNet)}
\label{alg:hqs}
\renewcommand{\algorithmicrequire}{\textbf{Input:}}
\renewcommand{\algorithmicensure}{\textbf{Output:}}
\begin{algorithmic}[1]
    \REQUIRE $\sigma_{0}$, $v$, $\mu > 0$, $z_{0}=\sigma_{0}$ 
    \ENSURE $\hat{\sigma}$    
    
    \FOR{$k=1, \cdots, K$}
    \STATE compute Jacobian matrix $J(\sigma)$ by \eqref{joca}
    \STATE compute $\sigma_{k} = \mathcal{G}(\sigma_{k-1};J,v,z_{k-1},\mu)$ with~\eqref{eq:hqs-net-gn}, details will be discussed in section \ref{GN}.
    \STATE compute $z_{k} = \Phi_{\theta}(\verb|c|(z_{k-1}, \frac{\partial \mathcal{L}}{\partial z_{k-1}}))$ with~\eqref{eq:hqs_net_prox}, details will be discussed in section \ref{lpgd}.
    \ENDFOR
    
    
    \RETURN $\hat{\sigma}=\sigma_{K}$
\end{algorithmic}
\end{spacing}
\end{algorithm}

Without unnecessary complications, we optimize the proposed HQSNet and AA-HQSNet methods using the widely adopted $\ell_{2}$ loss function. Given empirical data $(\sigma^{(n)}, v^{(n)})$ for $n=1,\cdots,N$, we denote the iterate in Algorithm~\ref{alg:hqs} by ${F}_N^{\dagger}(\sigma, \theta)$ and solve the following training process:
\begin{equation}\label{eq:mseloss}
\mathrm{L}(\theta):=\frac{1}{N} \sum_{n=1}^{N} \left\|{F}_N^{\dagger}\left(\hat{\sigma}^{(n)},\theta\right)-\sigma^{(n)}\right\|^{2}.
\end{equation}
Here, $\hat{\sigma}^{(n)}$ denotes the reconstruction result, $\theta$ represents the set of trainable parameters. After the learning process, the same algorithm scheme 
${F}_N^{\dagger}(\sigma, \theta^*$) with the learned parameters $\theta^*$ is subsequently utilized to make predictions on unseen data.

\subsubsection{Gauss-Newton Method for the First Sub-problem}\label{GN}
The Gauss-Newton method is an iterative optimization algorithm commonly used for solving nonlinear least squares problems.
It approximates the true Hessian matrix with a matrix calculated using the Jacobian matrix, making it computationally efficient and effective in solving nonlinear regression problems.
The Gauss-Newton method uses a line search strategy with a specific choice of decent direction. Given the non-linear least squares problem
 \begin{equation}\label{10}
 \sigma_{k}= \mathop{\arg\min}\limits_{\sigma}f(\sigma) \triangleq \mathop{\arg\min}\limits_{\sigma} \ \frac{1}{2} \Vert v - F(\sigma) \Vert^{2} + \frac{\mu}{2} \Vert z_{k-1} - \sigma \Vert^{2},
 \end{equation}
 and its first-order optimality prerequisite $\nabla f(\sigma^{*}) = 0$,
we apply the first-order Taylor's expansion approximation at $\sigma_{k}$:
\begin{equation*}\label{12}
\nabla f(\sigma^{*}) \approx \nabla f(\sigma_{k}) + \nabla^{2} f(\sigma_{k}) \Delta \sigma_{k} = 0,
\end{equation*}
where the gradient and the Hessian of $f(\sigma)$ are given respectively by 
\begin{equation*}
\nabla f(\sigma)  = J(\sigma)^{T}(F(\sigma)-v) + \mu (\sigma - z_{k-1}),
\end{equation*}
\begin{equation}\label{13}
\nabla^{2} f(\sigma) = J(\sigma)^{T} J(\sigma) + \sum_{k}r_{k}(\sigma) \nabla^{2}r_{k}(\sigma) + \mu I,
\end{equation}
with $J(\sigma)$ being the Jacobian matrix of $r(\sigma):=F(\sigma)-v$. The search direction $\Delta \sigma_{k}$ can be accessed by 
\begin{equation*}\label{15}
\Delta \sigma_{k} = - \nabla^{2}f(\sigma_{k})^{-1} \cdot \nabla f(\sigma_{k}).
\end{equation*}
The Gauss-Newton method essentially approximates the Hessian matrix by ignoring all the second-order terms that come from $\nabla^{2}f(\sigma)$ in \eqref{13}. The search direction $\Delta \sigma$ can be found by solving the linear system:
\begin{equation}\label{16}
(J^{T}(\sigma)J(\sigma) + \mu I) \Delta \sigma = - J^{T}(\sigma)(F(\sigma)-v) - \mu (\sigma - z_{k-1}).
\end{equation}
Starting from an initial guess $\sigma_{0}$, the Gauss-Newton algorithm (see Algorithm \ref{alg:1}) computes $\sigma_{k+1}$ based on the previous estimate $\sigma_{k}$ and the direction $\Delta \sigma_{k}$:
\begin{equation}\label{eq:hqs-net-gn}
\sigma_{k+1} = \mathcal{G}(\sigma_{k};J,v,z_{k},\mu) = \sigma_{k} + \Delta \sigma_{k}.
\end{equation}\par

\begin{algorithm}[htp]
\begin{spacing}{1.2}
\caption{Regularized Gauss-Newton Algorithm}
\label{alg:1}
\renewcommand{\algorithmicrequire}{\textbf{Input:}}
\renewcommand{\algorithmicensure}{\textbf{Output:}}
\begin{algorithmic}[1]
    \REQUIRE $\sigma_{0}$, $z_{0}$, $v$, $\mu > 0$  
    \ENSURE $\sigma_{k}$    
    
    \FOR{$k=1, \cdots, $}
    \STATE compute Jacobian matrix $J(\sigma)$
    \STATE compute direction $\Delta \sigma_{k}$ by (\ref{16})
    \STATE compute $\sigma_{k} = \sigma_{k-1} + \Delta \sigma_{k-1}$
    \ENDFOR
    
    
    \RETURN $\sigma_{k}$
\end{algorithmic}
\end{spacing}
\end{algorithm}
\subsubsection{Learned Proximal Gradient Network for the Second Sub-problem}\label{lpgd}
The proximal gradient operator algorithm for second sub-problem \eqref{9} contains two sub-problems: gradient descent~\eqref{eq:gd} and proximal mapping~\eqref{eq:prox}. It
starts from an initial value $z_0$ and performs the following iterates until convergence:
\begin{subequations}\label{eq:pgd}
\begin{align}
\hat{z}_{k}&=z_{k-1}-\alpha \frac{\partial \mathcal{L}}{\partial z_{k-1}}, \label{eq:gd}\\
z_{k} &= {\rm prox}_{R,\lambda}(\hat{z}_{k}), \label{eq:prox}
\end{align}
\end{subequations}
where $\mathcal{L} = \frac{\mu}{2}\Vert \sigma_{k}-z \Vert^{2}$, $\alpha$ is the step size, and the proximal operator $\mathcal{P}_{\lambda R}(\cdot)$ is defined  by
\begin{equation*}\label{prox}
{\rm prox}_{R,\lambda}(z)=
\mathop{\arg\min}\limits_{z} \frac{\mu}{2}\Vert \sigma_{k}-z \Vert^{2} + \lambda R(z).
\end{equation*}
Drawing inspiration from prior research \cite{xin2022learned}, we utilize a parameterized neural networks $\Phi_{\theta}$ to replace the proximal operators ${\rm prox}_{R,\lambda}$. Additionally, we concatenate the gradient term $\frac{\partial \mathcal{L}}{\partial z{k-1}}$ and the data term $z_{k-1}$ using the PyTorch function  \verb|torch.cat()|:
\begin{equation*}\label{prox4}
\hat{z}_{k}
=\verb|c| (z_{k-1},\frac{\partial \mathcal{L}}{\partial z_{k-1}}) 
= \verb|torch.cat(|z_{k-1},\frac{\partial \mathcal{L}}{\partial z_{k-1}}\verb|)|.
\end{equation*}
Subsequently, the learned proximal gradient descent algorithm here is to find a reconstruction operator defined as $\mathcal{F}_{\theta}^{\dagger}(z)=z_{t}$ where $z_{t}$ is given by the following finite recursive scheme initialised by $z_{0}\in X:$
\begin{equation}\label{eq:hqs_net_prox}
z_{k}=\Phi_{\theta}(\verb|c| (z_{k-1},\frac{\partial \mathcal{L}}{\partial z_{k-1}})).
\end{equation}

\subsection{Anderson Acceleration for Optimization Problems}\label{sec:aa}
Anderson acceleration (AA) is an effective method for accelerating the computation of fixed-point problems~\cite{anderson2019comments}. Here, we describe AA in the context of the classical gradient descent method used to minimize a smooth function $f$, as defined by
\begin{equation*}\label{321.1}
x_{k+1}=x_{k}- \gamma \nabla f(x_{k}),
\end{equation*}
which is equivalent to the fixed-point iteration applied to $g(x) = x - \gamma \nabla f(x)$. Notably, a fixed point $x^*$ of $g$ corresponds to an optimum of $f$, which enables the formulation of the solution to the optimization problem as $x^{*}=g(x^{*})$. 
In comparison to the fixed-point iteration, AA leverages past information more effectively. 
Given a sequence of estimates $\{x_{i}\}_{i=0}^{k}$ generated by AA up to iteration $k$, we refer to the residual in the $k$-th iteration as $r_{k}=g(x_{k})-x_{k}$. 
To determine $x_{k+1}$, AA searches for a point with the smallest residual within the subspace spanned by the $m+1$ most recent iterates.
In other word, if we let $\bar{x}_{k}=\sum_{i=0}^{m}\alpha_{i}^{k}x_{k-i}$, AA seeks to find a vector of coefficients $\alpha^{k}=[\alpha_{0}^{k}, \cdots, \alpha_{m}^{k}]^{\top}$ that satisfies the following optimization problem
\begin{equation}\label{aa}
\alpha^{k}=\mathop{\arg\min}\limits_{\alpha: \alpha^{\top}\textbf{1}=1}\Vert g(\bar{x}_{k}) -\bar{x}_{k} \Vert,
\end{equation}
where $\textbf{1}$ denotes the all-ones vector.
However, solving the optimization problem for a general non-linear mapping $g$ can pose significant difficulties. To overcome this challenge, AA utilizes the following expression:
\begin{equation}\label{aaa}
\alpha^{k}=\mathop{\arg\min}\limits_{\alpha:\alpha^{\top}\textbf{1}=1}\left\Vert \sum_{i=0}^{m}\alpha_{i}g(x_{k-i}) - \sum_{i=0}^{m}\alpha_{i}x_{k-i} \right\Vert.
\end{equation}
It is evident that problems \eqref{aa} and \eqref{aaa} are equivalent when $g_{1}$ represents an affine mapping. Let $R_{k}=[r_{k}, \cdots, r_{k-m}]$ denote the residual matrix at the $k$-th iteration. Problem \eqref{aaa} can then be rephrased as:
\begin{equation}\label{aaaa}
\alpha^{k}=\mathop{\arg\min}\limits_{\alpha:\alpha^{\top}\textbf{1}=1}\Vert R_{k}\alpha \Vert,
\end{equation}
once $\alpha^{k}$ is computed, the next iteration of AA is generated by
\begin{equation*}\label{}
x_{k+1}=\sum_{i=0}^{m}\alpha_{i}^{k}g(x_{k-i}).
\end{equation*}
When $m=0$, AA reduces to the fixed-point iteration.
When the optimization problem is solved using the Euclidean norm, the solution for $\alpha_k$ can be expressed as a closed-form equation given by
\begin{equation*}\label{}
\alpha^{k}=\frac{(R_{k}^{\top}R_{k})^{-1}\mathbf{1}}{\mathbf{1}^{\top}(R_{k}^{\top}R_{k})^{-1}\mathbf{1}},
\end{equation*}
this can be solved by first solving the $m\times m$ normal equations $R_{k}^{\top}R_{k}x=\mathbf{1}$ and then normalizing the result to obtain $\alpha^{k}=x/(\mathbf{1}^{\top}x)$.

\begin{algorithm}[htp]
\begin{spacing}{1.2}
\caption{Anderson Acceleration Algorithm}
\label{alg:aa}
\renewcommand{\algorithmicrequire}{\textbf{Input:}}
\renewcommand{\algorithmicensure}{\textbf{Output:}}
\begin{algorithmic}[1]
    \REQUIRE $x_{0}$, $g(\cdot)$, $m \geq 0$, 
    \ENSURE $x_{K}$    
    
    \STATE $x_{1} = g(x_{0})$   
    \FOR{$k=1, \cdots, K-1$}
    \STATE set $m_{k} = $ min$(m,k)$
    \STATE set $R_{k} = [r_{k},\cdots,r_{k-m_{k}}]$ with $r_{i}=g(x_{i})-x_{i}$
    \STATE solve $\alpha^{k}=\mathop{\arg\min}\Vert R_{k}\alpha \Vert$~~~s.t.~~${\alpha^{\top}\mathbf{1}=1}$
    \STATE set $x_{k+1}=\sum_{i=0}^{m_{k}}\alpha_{i}^{k}g(x_{k-i})$
    \ENDFOR

    \RETURN $x_{K}$
\end{algorithmic}
\end{spacing}
\end{algorithm}\par
The AA algorithm is outlined in Algorithm~\ref{alg:aa}. 
It is important to note that the number of past estimates $m$ dictates the amount of information from previous iterations utilized to compute the new velocity $x_t$. 
If $m$ is selected too small, insufficient information is used, potentially resulting in undesirably slow convergence. 
Conversely, if $m$ is excessively large, information from outdated iterations may be retained for excessive subsequent iterations, which may also lead to slow convergence. 
It is worth mentioning that the main purpose of this paper is to explore the application of AA technique to the deep unrolling models for nonlinear inverse problems.
In this context, a very interesting problem is developing
a tuning-free AA algorithm capable of automatically determining the optimal value of $m$.
In this paper, we set $m=2$ in accordance with ~\cite{evans2020proof, pollock2020benchmarking}, as  the authors suggested that a small value of $m$ would prove effective in practice.

\subsection{AA-HQSNet}\label{sec:hqs-aa-net}

\subsubsection{Gauss-Newton Anderson Acceleration}\label{sec:gnaa}
We address the optimization problem \eqref{10} using the Newton method, which updates $\sigma$ iteratively as follows:
\begin{equation*}\label{322.2}
\sigma_{k+1}=\sigma_{k}- \beta [\nabla^{2} f(\sigma_{k})]^{-1} \cdot \nabla f(\sigma_{k}),
\end{equation*}
where $\beta \in (0,1]$.
The work of \cite{pollock2020benchmarking} verifies the theoretical and numerical benefits of the Newton Anderson acceleration (Newton-AA) for enhancing the local convergence rate of linear optimization problems. However, the computational complexity associated with the Hessian matrix in the Newton-AA method raises concerns. To tackle this issue, we introduce a novel Gauss-Newton method combined Anderson acceleration (Gauss-Newton-AA). We introduce the following notation. The Gauss-Newton iteration can be expressed as:
$\sigma_{k+1}=g(\sigma_{k})=\sigma_{k}+(\phi (\sigma_{k})-\sigma_{k})=\sigma_{k}+\beta w_{k+1}$,
where $w_{k+1}=-[\nabla^{2}f(\sigma_{k})]^{-1}\cdot \nabla f(\sigma_{k})$. Note that the Hessian matrix  $\nabla^{2} f(\sigma)$ is replaced by the Jacobian approximation: 
\begin{equation}\label{322.6}
H(\sigma)=\nabla^{2} f(\sigma) \approx \nabla f(\sigma)^{T}\cdot \nabla f(\sigma)=\tilde{H}(\sigma).
\end{equation}
Clearly, the Gauss-Newton iteration aligns with the fixed point iteration scheme presented in Section \ref{sec:aa}, ensuring that the AA method remains applicable.
The Gauss-Newton-AA approach aims to find a vector of coefficients $\gamma^{k+1}=[\gamma_{0}^{k}, \gamma_{1}^{k}, \cdots, \gamma_{m}^{k}]^{\top}$ that satisfies the following equation 
by using the notation $\sigma_{k+1}=g(\sigma_{k})=\sigma_{k}+ \beta w_{k+1}$ where $w_{k+1}=-(\tilde{H}(\sigma))^{-1} \cdot \nabla f(\sigma_k)$

\begin{equation}\label{gnaa1}
    \gamma^{k+1} = \mathop{\arg\min}\limits_{\gamma \in \mathbb{R}^m} \left\Vert w_{k+1} - \sum_{i=0}^m \gamma_i \Delta w_{k-i+1} \right\Vert,
\end{equation}
where $\Delta w_{k-i+1}$ is defined as:
\begin{equation*}
    \Delta w_i = w_{i+1} - w_i, \quad i = k, \ldots, k-m+1,
\end{equation*}
the $\alpha$ defined in~\eqref{aaaa} and $\gamma$ are related by~\cite{pollock2021anderson}
\begin{equation*}
\alpha_i = 
\left\{\begin{array}{ll}
1 - \gamma_0 &\quad i=0, \\ 
\gamma_{i-1} - \gamma_i &\quad 0<i<m,\\ 
\gamma_m &\quad i=m,
\end{array}\right.
\end{equation*}
using $F_{k}=[(w_{k+1}-w_{k}), \cdots, (w_{k-m+2}-w_{k-m+1})]$, problem \eqref{gnaa1} can be expressed as
\begin{equation*}
    \gamma_{k+1}=\mathop{\arg\min}\limits_{\gamma \in R^{m}} \Vert w_{k+1}-F_{k}\gamma \Vert,
\end{equation*}
once $\gamma^{k+1}$ is computed, $\Bar{\sigma}_k=\sum_{i=0}^m \alpha_i^k \sigma_{k-i}$ as defined in Section~\ref{sec:aa} is known, and the next iteration of Gauss-Newton-AA is generated as follows:
\begin{equation*}        
    \sigma_{k+1}=g(\bar{\sigma}_{k}) = \sigma_{k} + \beta w_{k+1} - (E_{k}+\beta F_{k})\gamma^{k+1},
\end{equation*}
where $E_{k}=[(\sigma_{k}-\sigma_{k-1}), \cdots, (\sigma_{k-m+1}-\sigma_{k-m})]$.
After $K$ iterations of the above algorithm, the final result of the Gauss-Newton-AA method can be expressed as:
\begin{equation}\label{eq:GNAA}
    \sigma_{K}=\mathcal{G}^{AA}(\sigma_{0}; z_{0}, J,v,\mu, \beta, m, K),
\end{equation}
where $\mathcal{G}^{AA}$ represents the aforementioned procedure with a total of $K$ iterations. 
In our AA-HQSNet, during the $i$-th iteration, we assume that $\sigma^{K^{(1)}}_{i-1}$ and $z^{K^{(2)}}_{i-1}$ are already known. Here, the subscript $i$ indicates the number of alternating iterative updates of $\sigma$ and $z$, while the superscript $K^{(1)}$ represents the total number of iterations of Gauss-Newton AA. Similarly, the superscript $K^{(2)}$ represents the iterations of Anderson acceleration of the learned proximal gradient descent (AA-LPGD) method, which we will discuss in the next section. We can restate Eq.\eqref{eq:GNAA} by defining $\sigma^{0}_{i}=\sigma^{K^{(1)}}_{i-1}$ and $z^{0}_{i}=z^{K^{(2)}}_{i-1}$, as follows:
\begin{equation}\label{eq:hqs-aa_net-sigma}
\sigma_{i}=\sigma^{K^{(1)}}_{i}=\mathcal{G}^{AA}(\sigma^{0}_{i}; z^{0}_{i}, J,v,\mu, \beta, m^{(1)}, K^{(1)}),
\end{equation}
The overall methodology of the Gauss-Newton-AA method in the $i$-th iteration of AA-HQSNet is presented in Algorithm \ref{alg:gnaa} and a graphical illustration of the architecture can be found in Fig.\ref{fig:hqsaa_str}\textbf{(b)}.
\begin{algorithm}[htp]
\begin{spacing}{1.2}
\caption{Gauss-Newton Anderson Acceleration: $ \sigma^{K}_{i} \gets \mathcal{G}^{AA}(\sigma^{0}_{i}; z^{0}_{i}, J,v,\mu, \beta, m, K)$}
\label{alg:gnaa}
\renewcommand{\algorithmicrequire}{\textbf{Input:}}
\renewcommand{\algorithmicensure}{\textbf{Output:}}
\begin{algorithmic}[1]
    \REQUIRE $\sigma^{0}_{i}$, $\beta \in (0,1]$, $m=m^{(1)} \geq 0$, $K=K^{(1)} \geq 1$, $v$, $z^{0}_{i}$, $\mu > 0$ 
    \ENSURE $\sigma^{K^{(1)}}_{i}$    
    
    \STATE compute $w_{1}=-(\tilde{H}(\sigma^{0}_{i}))^{-1} \cdot \nabla f(\sigma_i^0)$ by \eqref{322.6}
    \STATE set $\sigma^{1}_{i}=\sigma^{0}_{i}+\beta w_{1}$
    \FOR{$k=1, \cdots, K^{(1)}-1$}
    \STATE $m_{k} = $ min$(m^{(1)},k)$
    \STATE compute $w_{k+1}=-(\tilde{H}(\sigma^{k}_{i}))^{-1} \cdot \nabla f(\sigma_i^k)$ by \eqref{322.6}
    \STATE set $F_{k}=[(w_{k+1}-w_{k}),\cdots, (w_{k-m_{k}+2}-w_{k-m_{k}+1})]$, and $E_{k}=[(\sigma^{k}_{i}-\sigma^{k-1}_{i}),\cdots, (\sigma^{k-m_{k}+1}_{i}-\sigma^{k-m_{k}}_{i})]$
    \STATE compute $\gamma^{k+1}=\mathop{\arg\min}\limits_{\gamma \in \mathbb{R}^{m^{(1)}}} \Vert w_{k+1}-F_{k}\gamma \Vert$
    \STATE set $\sigma^{k+1}_{i}=\sigma^{k}_{i}+\beta w_{k+1}-(E_{k}+\beta F_{k})\gamma^{k+1}$
    \ENDFOR
    
    
    \RETURN $\sigma^{K^{(1)}}_{i}$
\end{algorithmic}
\end{spacing}
\end{algorithm}\par
Despite using the Jacobian approximation for the Hessian matrix, the Gauss-Newton-AA method maintains comparable effectiveness to the Newton-AA method, as shown in Example 3.3.1.\par
\begin{figure}[htp]
\centering
\subfigure{
\begin{minipage}[b]{.45\linewidth}
\centering
\includegraphics[scale=0.055]{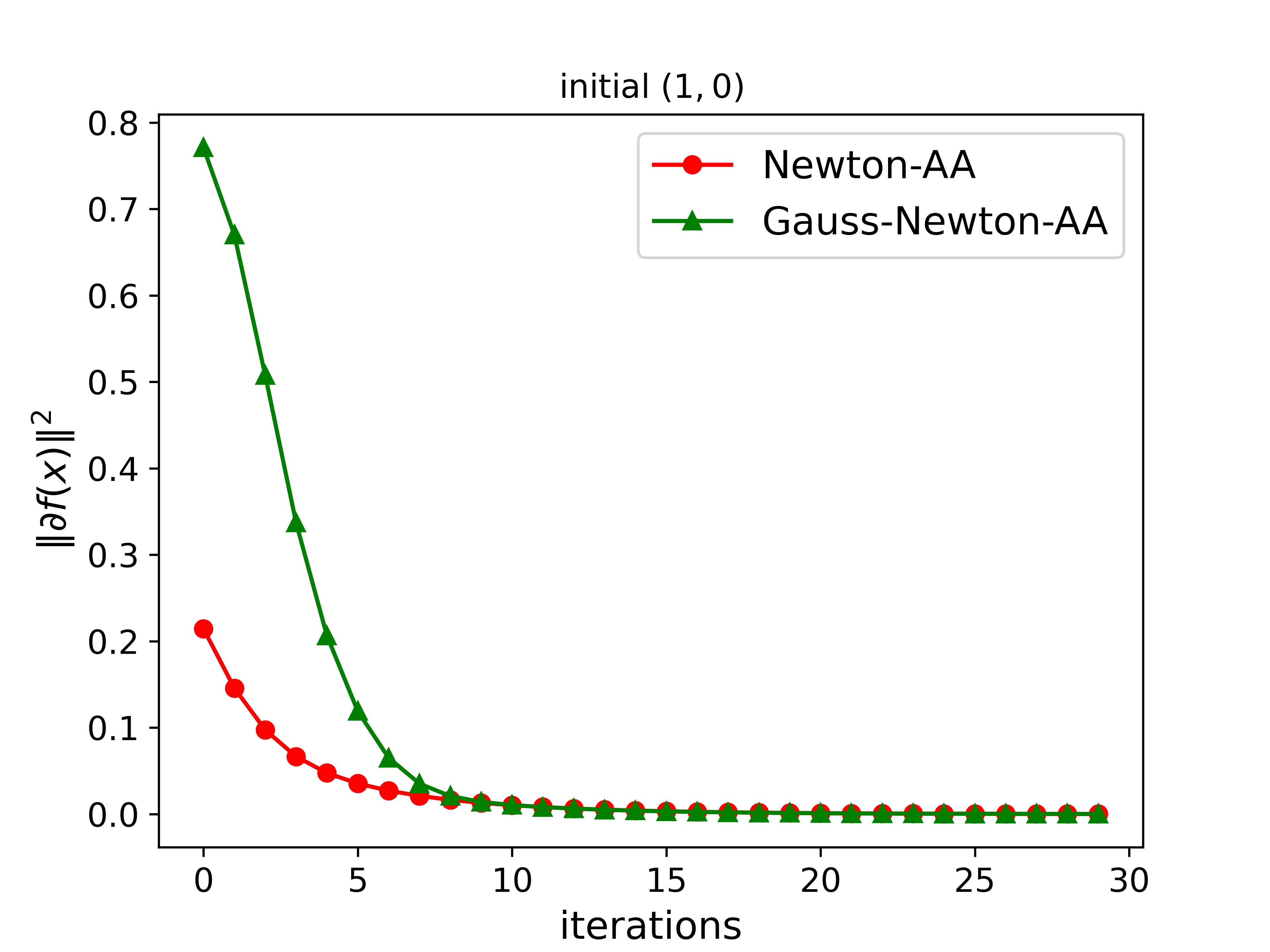}
\end{minipage}
}
\subfigure{
\begin{minipage}[b]{.45\linewidth}
\centering
\includegraphics[scale=0.055]{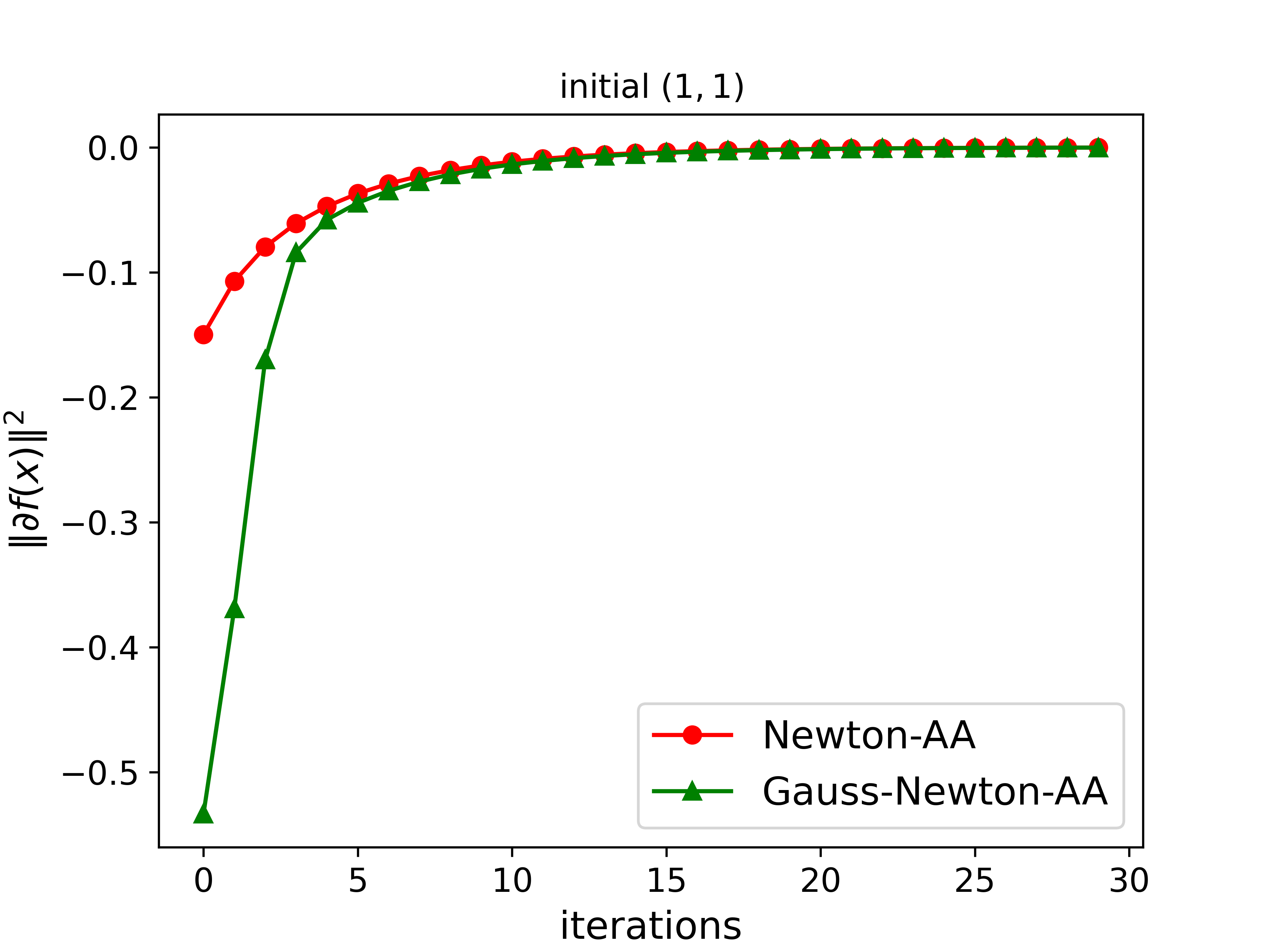}
\end{minipage}
}
\subfigure{
\begin{minipage}[b]{.45\linewidth}
\centering
\includegraphics[scale=0.055]{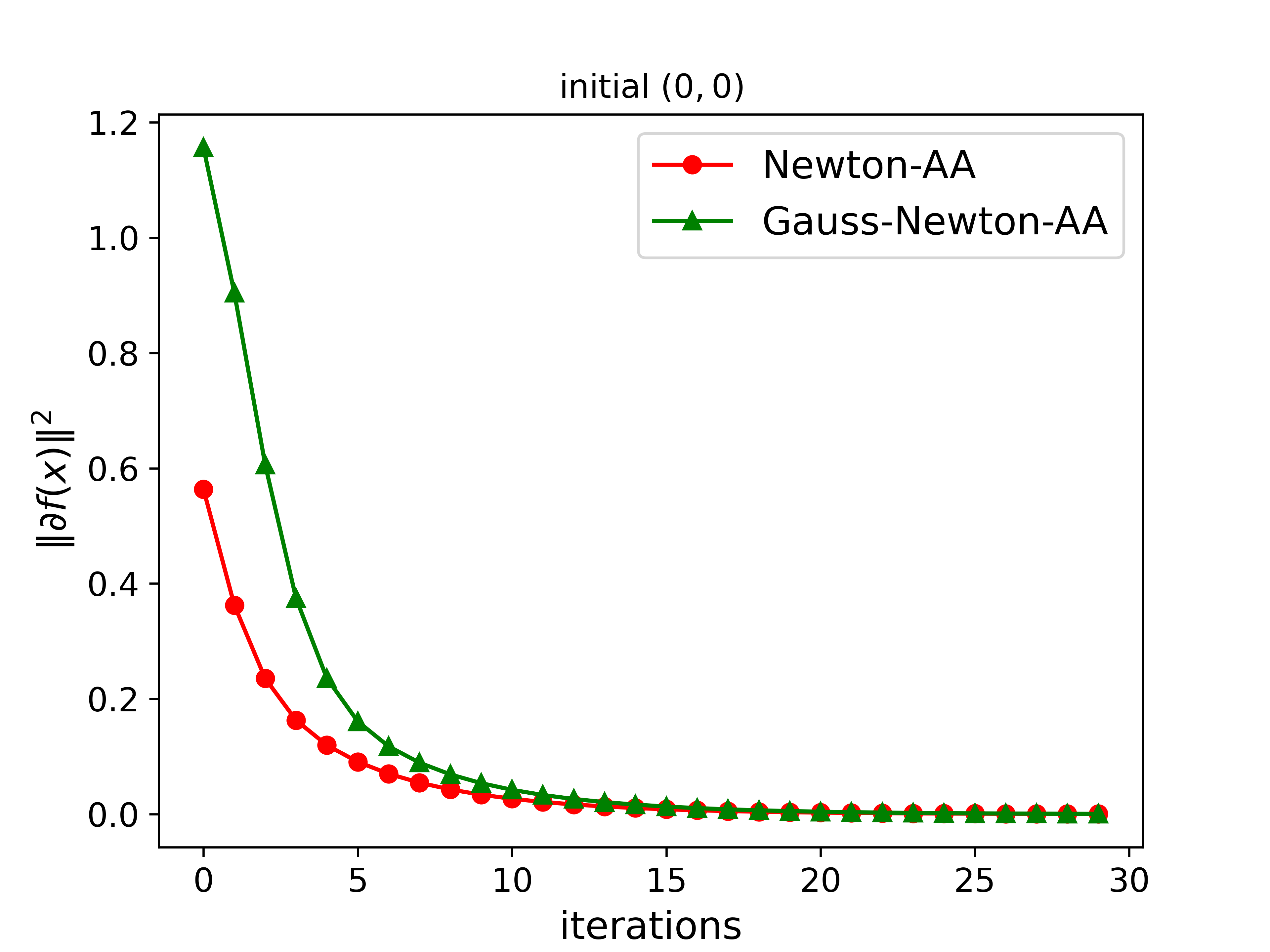}
\end{minipage}
}
\subfigure{
\begin{minipage}[b]{.45\linewidth}
\centering
\includegraphics[scale=0.055]{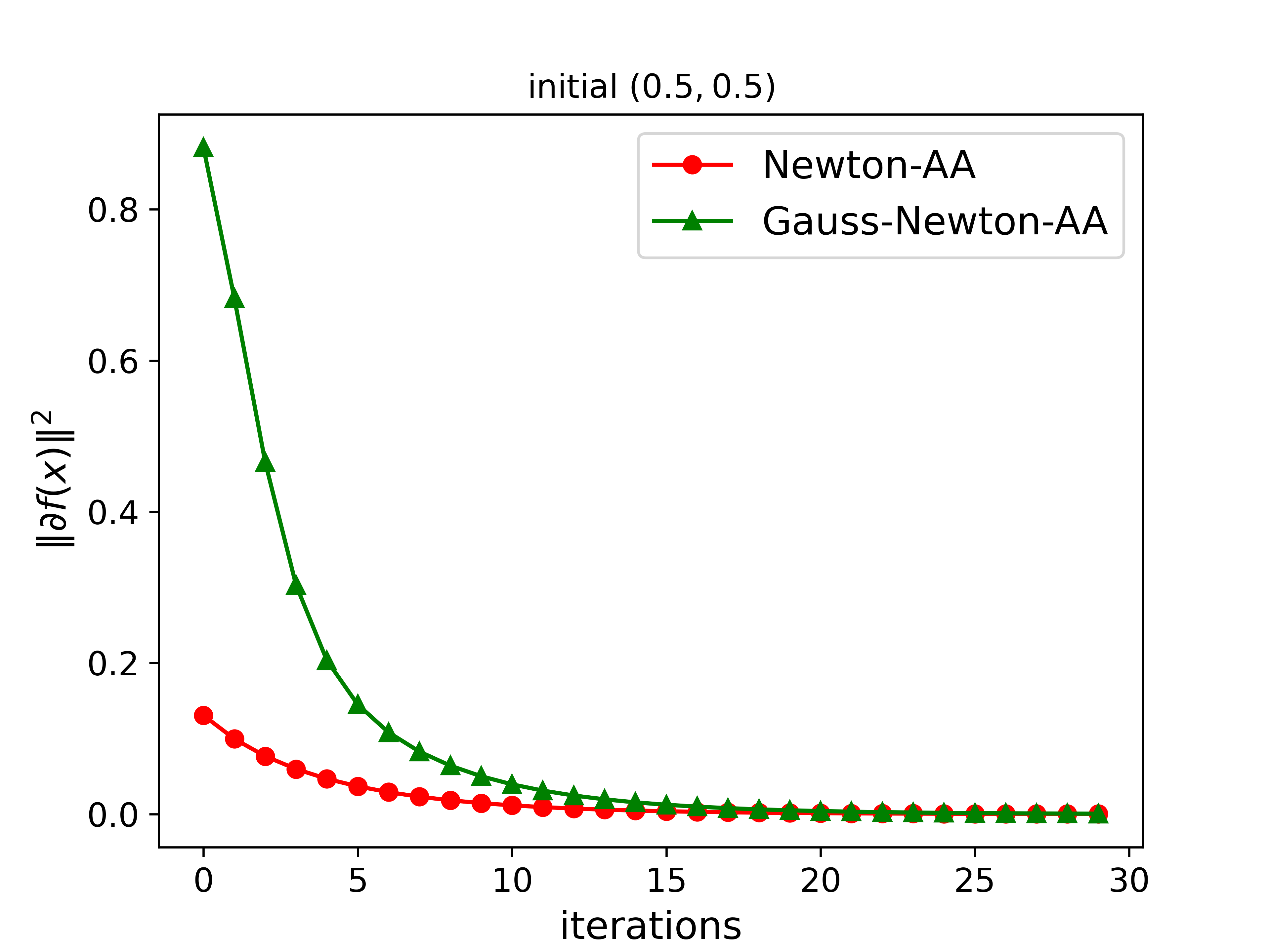}
\end{minipage}
}
\caption{Comparison of performance between Newton-AA (NAA) and Gauss-Newton-AA (GNAA) with different initial values, showcasing the variation of the $l_{2}$ norm of $f(x)$ (y-axis) across iterations (x-axis)}
\label{fig:gnaa}
\end{figure}

\textbf{Example 3.3.1.} 
\textit{
As an illustrative example highlighting the contrast between the Newton-AA and the Gauss-Newton-AA techniques, we consider a simple optimization problem with a function  $f:\mathbb{R}^{2}\to \mathbb{R}^{2}$ introduced in \cite{pollock2020benchmarking}, which can be defined as follows:
\begin{equation*}
f(x)=\left(\begin{array}{c}
10^4 x_1 x_2-1 \\
\exp \left(-x_1\right)+\exp \left(-x_2\right)-1.0001
\end{array}\right)
\end{equation*}
with different initial $x_{0}$. To begin with, we solve this optimization problem using Newton-AA with $m=\{1, 2, 5, 10, 15\}$. Here, $\nabla^{2}f(x)$ can be calculated through a precise mathematical derivation. Alternatively, we approximate $\nabla^{2}f(x)$ using the Gauss-Newton method: $\nabla^{2} f(x) \approx \nabla f(x)^{T}\cdot \nabla f(x)$. The results, as depicted in Fig.\ref{fig:gnaa}, indicate that the convergence of Gauss-Newton-AA is initially inferior to that of Newton-AA. However, with a suitable initial point and the choice of $m$, their convergence abilities can become comparable, even better.}

\subsubsection{Anderson Acceleration of Learned Proximal Gradient Descent}\label{sec:pgdaa}
We improve the performance of the learned proximal gradient descent (LPGD) method, presented in Section \ref{lpgd}, by incorporating the Anderson acceleration module, resulting in the development of the Anderson accelerated learned proximal gradient descent (AA-LPGD) method.
Specifically, sub-problem \eqref{9} can be solved through an iterative update of $z_{k}=\Phi_{\theta}(\verb|c|(z_{k-1},\frac{\partial \mathcal{L}}{\partial z_{k-1}}))$, which can be seen as a fixed-point iteration
\begin{equation}\label{332.1}
z^{*}=\Phi_{\theta}(\verb|c|(z^{*}, \nabla \mathcal{L}(z^{*}))),
\end{equation}
wherein $z^{*}$ represents the minimizer of \eqref{9}. To expedite this iterative scheme, we reformulate \eqref{332.1} as follows
\begin{equation}\label{332.2}
y_{k+1}=\verb|c|(z_{k}, \nabla \mathcal{L}(z_{k})), \ \text{and} \ z_{k+1}=\Phi_{\theta}(y_{k+1}),
\end{equation}
and define the mapping $g$ as 
\begin{equation}\label{332.3}
g(y)=\verb|c|(\Phi_{\theta}(y), \nabla \mathcal{L}(\Phi_{\theta}(y))).
\end{equation}
Hence, we can represent the proximal gradient iteration \eqref{332.2} as
\begin{equation}\label{332.4}
    y_{k+1}=g(y_{k}).
\end{equation}
Given that $y^{*}$ is a fixed point of $g$, which can be expressed as:
\begin{equation*}
    y^{*}=g(y^{*})=\verb|c|(z^{*}, \nabla \mathcal{L}(z^{*})),
\end{equation*}
then $z^{*}=\Phi_{\theta}(y^{*})$ is also a fixed point of $\Phi_{\theta}$, as indicated in \eqref{332.1}. This implies that $z^{*}$ is an optimal solution to \eqref{9}. To expedite the convergence of the auxiliary sequence $\left\{y_{k}\right\}$ governed by $g$ as defined in \eqref{332.3}, the Anderson acceleration method can be employed. Following the idea of generic Anderson acceleration, let $r_{k}=g(y_{k})-y_{k}$ and $\bar{y}_{k}=\sum_{i=0}^{m}\alpha_{i}^{k}y_{k-i}$, then $\alpha^{k}=[\alpha_{0}^{k}, \cdots, \alpha_{m}^{k}]^{\top}$ can be obtained by
\begin{equation*}
        \alpha^{k} = \mathop{\arg\min}\limits_{\alpha: \alpha^{\top}\textbf{1}=1} \Vert g(\bar{y}_{k})-\bar{y}_{k} \Vert 
        = \mathop{\arg\min}\limits_{\alpha: \alpha^{\top}\textbf{1}=1} \Vert R_{k}\alpha \Vert,
\end{equation*}
with $R_{k}=[r_{k}, \cdots, r_{k-m}]$, the next iteration of AA is then computed as
\begin{equation*}
    y_{k+1}=g(y_{k})=\sum_{i=0}^{m}\alpha_{i}^{k}g_{k-i},
\end{equation*}
and
\begin{equation*}
    z_{k+1}=\Phi_{\theta}(y_{k+1}),
\end{equation*}
according to Eq.\eqref{332.1}.
We denote the output after $K$ iterations as 
\begin{equation}\label{eq:AAPGA}
    z_{K}=\Phi^{AA}_{\theta}(z_{0}; \sigma_{0}, v, \mu, m, K).
\end{equation}
Specifically, within our AA-HQSNet, in the $i$-th iteration, given that $\sigma^{K^{(1)}}_{i}$ and $z^{K^{(2)}}_{i-1}$ are known, in which the subscript $i$ denotes the number of alternating iterative updates of $\sigma$ and $z$, the superscript $K^{(2)}$ represents a total number of iterations of AA-LPGD and the superscript $K^{(1)}$ represents the iterations of Gauss-Newton-AA method discussed in Section \ref{sec:gnaa}. By the notion of $\sigma^{0}_{i+1}=\sigma^{K^{(1)}}_{i}$ and $z^{0}_{i}=z^{K^{(2)}}_{i-1}$, we can rewrite Eq.\eqref{eq:AAPGA} as
\begin{equation}\label{eq:hqs-aa_net-z}
       z_{i}=z^{K^{(2)}}_{i}=\Phi_{\theta}^{AA}(z^{0}_{i}; \sigma^{0}_{i+1}, v, \mu, m^{(2)}, K^{(2)}),
\end{equation}
we summarize AA-LPGD within $i$-th iteration of AA-HQSNet in Algorithm \ref{alg:pgdaa}, and the structure of AA-LPGD is shown in Fig.\ref{fig:hqsaa_str}\textbf{(c)}.
\begin{algorithm}[htp]
\begin{spacing}{1.2}
\caption{AA-Learned Proximal Gradient Descent Algorithm: $z^{K}_{i}\gets \Phi^{AA}_{\theta}(z^{0}_{i}; \sigma^{0}_{i+1}, v, \mu, m, K)$}
\label{alg:pgdaa}
\renewcommand{\algorithmicrequire}{\textbf{Input:}}
\renewcommand{\algorithmicensure}{\textbf{Output:}}
\begin{algorithmic}[1]
    \REQUIRE $z^{0}_{i}$, $m=m^{(2)} \geq 0$, $K=K^{(2)} \geq 1$, $\sigma^{0}_{i+1}$, $v$, $\mu$ 
    \ENSURE $z^{K^{(2)}}_{i}$    
    
    \STATE $y_{0}=z^{0}_{i}$ 
    \STATE $y_{1}= \verb|c|(z^{0}_{i}, \nabla \mathcal{L}(z^{0}_{i}))$, $z^{1}_{i} = \Phi_{\theta}(y_{1})$, $g_{0} = y_{1}$
    \FOR{$k=1, \cdots, K^{(2)}-1$}
    \STATE $m_{k} =$ min$(m^{(2)},k)$
    \STATE $g_{k}= \verb|c|(z^{k}_{i}, \nabla \mathcal{L}(z^{k}_{i}))$ and $r_{k}= g_{k}-y_{k}$
    \STATE $R_{k}= [r_{k}, \cdots, r_{k-m_{k}}]$
    \STATE $\alpha^{k}= \mathop{\arg\min}\limits_{\alpha^{\top}\mathbf{1}=1}\Vert R_{k}\alpha \Vert$
    \STATE $y_{k+1}= \sum_{i=0}^{m_{k}}\alpha_{i}^{k}g_{k-i}$
    \STATE $z^{k+1}_{i}= \Phi_{\theta}(y_{k+1})$
    \ENDFOR
    
    \RETURN $z^{K^{(2)}}_{i}$
\end{algorithmic}
\end{spacing}
\end{algorithm}\par

\subsubsection{Summary of AA-HQSNet Algorithm}
\begin{figure}[htp]
\centering
\includegraphics[width=1\textwidth]{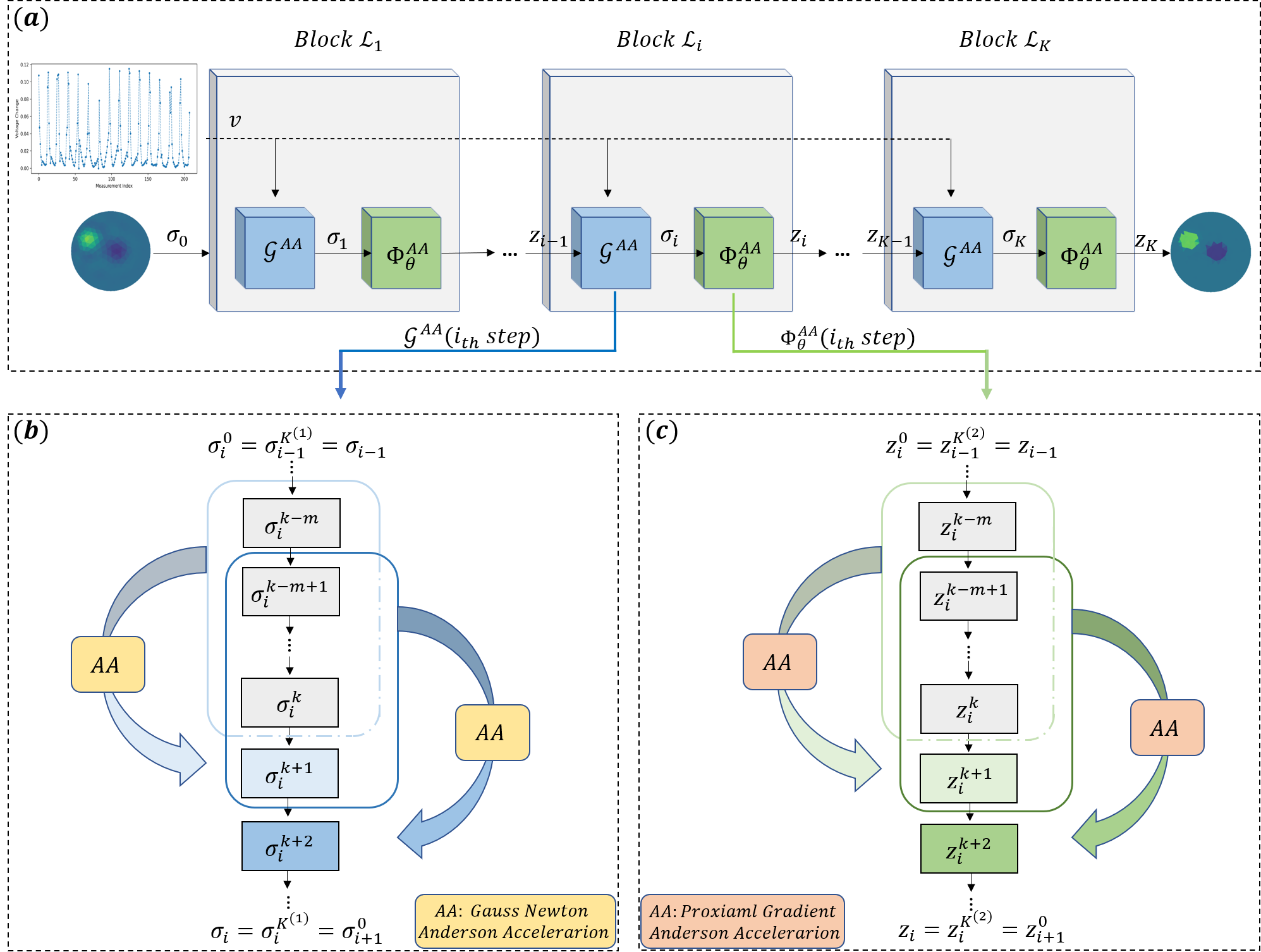}
\caption{\textbf{(a)}The unrolled AA-HQSNet architecture, where $\mathcal{G}$ represents the Gauss-Newton method summarized in Algorithm \ref{alg:1}, and $\Phi_{\theta}$ represents the proximal neural network depicted in Fig.\ref{fig:network}. \textbf{(b)} The Gauss-Newton Anderson acceleration algorithm with iteration steps $K^{(1)}$, as shown in Algorithm \ref{alg:gnaa}. \textbf{(c)} The proximal gradient descent Anderson acceleration algorithm with iteration steps $K^{(2)}$, introduced in Algorithm \ref{alg:pgdaa}}
\label{fig:hqsaa_str}
\end{figure}

Now we outline the complete scheme for the AA-HQSNet algorithm using the methods discussed in sections~\ref{sec:gnaa} and~\ref{sec:pgdaa}. 
By integrating the AA technique into our HQSNet, we can iteratively solve optimization problem \eqref{6} via Eq.\eqref{eq:hqs-aa_net-sigma} and Eq.\eqref{eq:hqs-aa_net-z}. 
With a predetermined number of iterations $K$, $K^{(1)}$, $K^{(2)}$, the AA-HQSNet initially accelerates $\sigma$ and $z$ iteratively within Gauss-Newton-AA step and the learned proximal-gradient step, respectively. 
Subsequently, we alternately update $\sigma$ and $z$ externally. The complete AA-HQSNet algorithm is summarized in Algorithm \ref{alg:hqsaa}, and the graphical structure is depicted in Fig.\ref{fig:hqsaa_str}.

\begin{algorithm}[htp]
\begin{spacing}{1.2}
\caption{Learned Half-Quadratic Splitting with Anderson Acceleration (AA-HQSNet)}
\label{alg:hqsaa}
\renewcommand{\algorithmicrequire}{\textbf{Input:}}
\renewcommand{\algorithmicensure}{\textbf{Output:}}
\begin{algorithmic}[1]

    \REQUIRE \textit{General parameters}: $K$, $v$, $\mu > 0$;\\ \ \ \ \ \ \
    \textit{Gauss-Newton-AA's parameters}: $\sigma^{0}_{1}=\sigma_{0}$, $K^{(1)}$, $m^{(1)}$, $\beta \in (0,1]$;\\ \ \ \ \ \ \ \textit{AA-LPGD's parameters}: $z^{0}_{1}=z_{0}$, $K^{(2)}$, $m^{(2)}$

    \ENSURE $\hat{\sigma}$    

   \FOR{$i=1, \cdots, K$}
    \STATE compute $    \sigma_{i}=\sigma^{K^{(1)}}_{i}=\mathcal{G}^{AA}(\sigma^{0}_{i}; z^{0}_{i}, J,v,\mu, \beta, m^{(1)}, K^{(1)})$ with Eq.\eqref{eq:hqs-aa_net-sigma}
    \STATE set $\sigma^{0}_{i+1}=\sigma^{K^{(1)}}_{i}$
    \vspace{7pt}
    \STATE compute $       z_{i}=z^{K^{(2)}}_{i}=\Phi_{\theta}^{AA}(z^{0}_{i}; \sigma^{0}_{i+1}, v, \mu, m^{(2)}, K^{(2)})$ with Eq.\eqref{eq:hqs-aa_net-z}
    \STATE set $z^{0}_{i+1}=z^{K^{(2)}}_{i}$
    \ENDFOR
    
    \RETURN $\hat{\sigma}=\sigma_{K}$
\end{algorithmic}
\end{spacing}
\end{algorithm}

\par

\section{Experimental Results}\label{sec:experiment}
\subsection{Implementation Details}\label{ID}
We employ the experimental setup presented in~\cite{colibazzi2022learning} to generate our simulated datasets.
The simulations are designed with circular meshes consisting of 660 triangles and a circular structure containing 16 evenly distributed electrodes along the circular boundary ring. The conductivity of the background medium is set to $\sigma_{0}=1.0 \ {\rm \Omega} m^{-1}$. To simulate measurements, we employ the \textit{opposite injection-adjacent measurement} protocol via pyEIT \cite{LIU2018304}, a Python-based framework for electrical impedance tomography. During the experiment, no prior knowledge of the settings is available, i.e., there is no prior information about the sizes or locations of the inclusions. The training process utilized a total of 250 randomly generated samples, comprising 200 training samples and 50 test samples.  The complete training dataset contained 200 pairs of real conductivity values $\sigma^{GT}$ and the corresponding measured voltages $v$. Each sample consists of a circular tank containing a random number of anomalies ranging from 1 to 4 with random locations and characterized by random radii within the range $[0.15, 0.25]$ and magnitudes in the range $[0.2,2]$.\par
\begin{figure}[htp]
    \centering
    \includegraphics[width=0.8\textwidth]{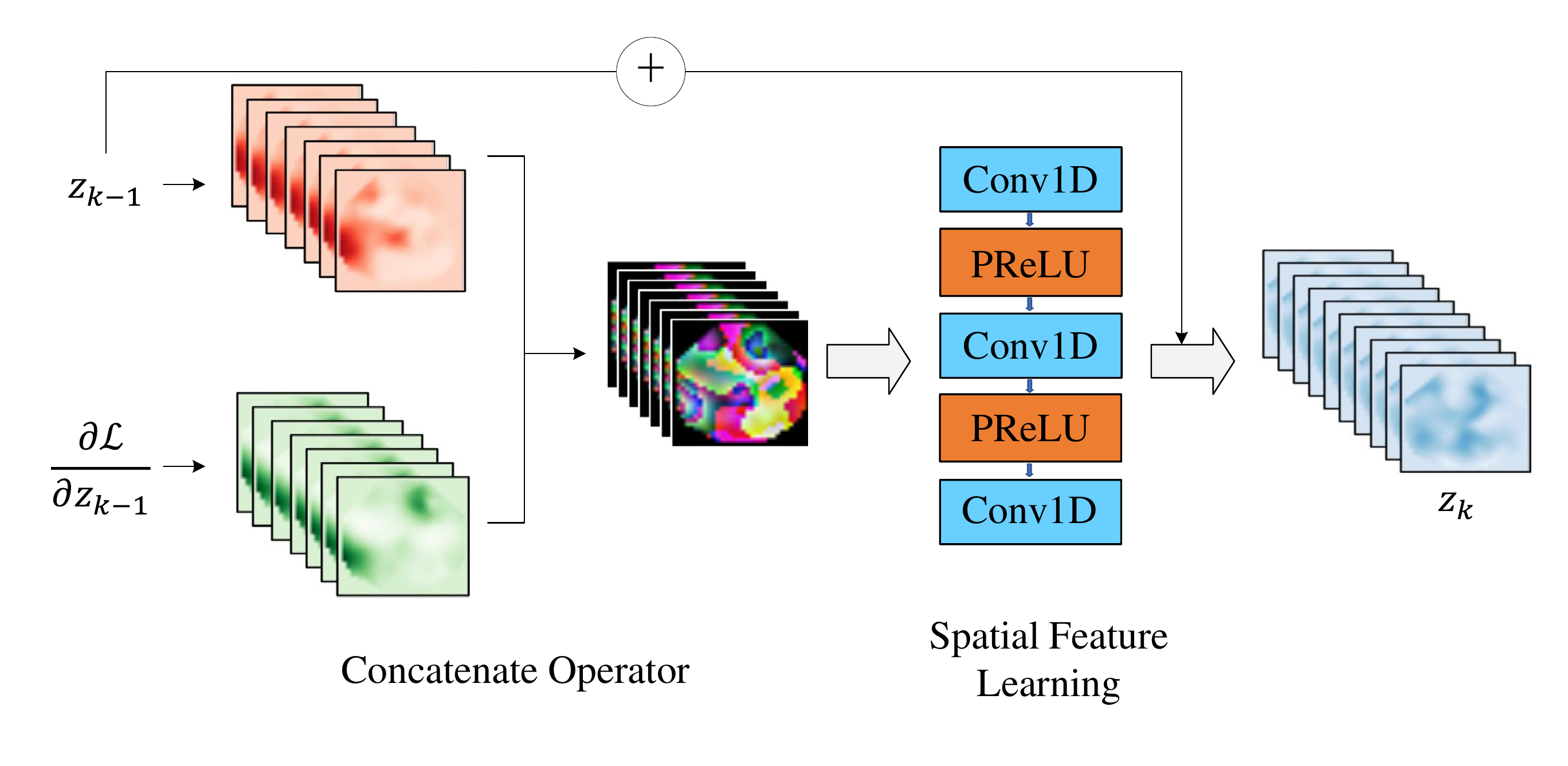}
    \caption{Unrolled proximal gradient network architecture $\Phi_{\theta}$ used in proximal-gradient decent steps, and its input is the data term and corresponding gradient term}
    \label{fig:network}
\end{figure}

The proximal network used in the AA-HQSNet follows the structure depicted in Fig.\ref{fig:network}. The network comprises a concatenate layer that combines the data term and the gradient term, followed by several convolutional layers (with 3 $\times$ 3 kernels and 32 feature maps) and a PReLU activation function.\par
\par

The learning rate is set to $10^{-3}$ and learned iteratively over 30 epochs. Each epoch takes approximately 6 minutes. 
All experiments are implemented in Python and executed on a laptop equipped with eight 3.20 GHz cores and 16 GB of RAM, running Windows \textit{10.0}. The implementation of all experiments is accessible at \href{https://github.com/CCcodecod/AA-HQSNet}{https://github.com/CCcodecod/AA-HQSNet}.

\subsection{Metrics}\label{metric}
Experimental performance is evaluated using both qualitative and quantitative metrics. The quantitative analysis is based on mesh elements and is defined by the mean squared error (MSE) as follows:
\begin{equation*}
MSE = \frac{\Vert \sigma^{GT} - \hat{\sigma} \Vert_{2}^{2}}{n_{T}},
\end{equation*}
where $\sigma^{GT}$ is the ground truth conductivity distribution and $\hat{\sigma}$ is the reconstructed conductivity distribution.  The MSE measures the degree of accuracy in reconstructing the original conductivity distribution.
In addition to the MSE, the similarity between two grided images is also measured using the Structural SIMilarity (SSIM) Index, which is modified to work directly on the mesh \cite{colibazzi2022learning}. In the image field, $SSIM(A,B)$ quantifies the inconsistency between A and B, with $SSIM=1$ indicating that A and B are equivalent, and $SSIM=0$ indicating that the two images are completely different.\par
Drawing from the work of Colibazzi et al. \cite{colibazzi2022learning}, we use a new assessment index, termed Electrical Impedance Tomography Evaluation Index (EIEI), to evaluate the homogeneity of anomalies in EIT.
The number of triangles occupied by backgrounds, artifacts and anomalies are denoted as $n_{1}$, $n_{2}$, and $n_{3}$, respectively. The associated conductivity values are denoted by $\sigma^{(1)}$, $\sigma^{(2)}$, and $\sigma^{(3)}$.  In contrast to \cite{8362785}, where all anomalies are assumed to have the same values, we adopt the clustering method to estimate the variances of these clusters, as described in \cite{colibazzi2022learning}:
\begin{equation*}
\delta_{j}=\sum_{i=1}^{n_{j}}\vert \sigma_{i}^{(j)}-\bar{\sigma}^{(j)} \vert/n_{j}, \ \ \ j=1,2,3.
\end{equation*}
where $\sigma_{i}^{(j)}$ represents the conductivity value associated with the $i_{th}$ triangle with a value of $\sigma^{(j)}$, and $\bar{\sigma}^{(j)}$ denotes the average conductivity value. The EIEI metric is then defined as follows:
\begin{equation}\label{eieimetric}
EIEI=w_{1}T_{1}+w_{2}T_{2},
\end{equation}
with $T_{1}=1-n_{2}/n_{T}$ being the number of artifact triangles and $T_{2}=1-(\delta_{1}n_{1}/n_{T}+\delta_{3}n_{3}/n_{T})$ being the number of backgrounds and anomalies triangles. The weighting values $w_{1}$ and $w_{2}$ represent the certainty of $T_{1}$ and $T_{2}$, and are evaluated as
\begin{equation*}
w_{1}=\sum_{i=1}^{n_{2}}\sigma_{i}^{(2)}/n_{2}, \ \ w_{2}=\left(\sum_{i=1}^{n_{1}}\sigma_{i}^{(1)}+\sum_{i=1}^{n_{3}}\sigma_{i}^{(3)}\right)/(n_{1}+n_{3}).
\end{equation*}
This metric provides qualitative insight via an EIEI map, where yellow indicates artifacts, red indicates anomalies, and blue represents background.\par
Dynamic range is another important metric employed in this study to assess the quality of the reconstructed conductivity distribution, which is defined as
\begin{equation*}
DR=\frac{\max \hat{\sigma}- \min \hat{\sigma}}{\max \sigma^{GT}-\min \sigma^{GT}} \times 100\%,
\end{equation*}
a DR value closer to 100 signifies optimal reconstruction quality.

\subsection{Performance Comparisons}\label{comparisons}
We conduct a comparative analysis of our proposed HQSNet and AA-HQSNet methods against traditional methods such as D-bar~\cite{doi:10.1137/1.9781611972344} and Gauss-Newton with the Levemberg-Marquardt (GN-LM)~\cite{LIU2018304}, as well as the state-of-the-art deep unrolling method, EITGN-NET \cite{colibazzi2022learning}. 
To ensure a fair comparison, all parameters involved in the D-bar and GN-LM methods are either manually tuned optimally or automatically selected as described in the reference papers.

Table \ref{tab:per_comp} presents a comparison of the performance of all methods on the testing dataset. It is evident that both AA-HQSNet and HQSNet algorithms significantly outperform the state-of-the-art EINGN-NET model. For example, AA-HQSNet demonstrates a 17.2\% reduction in MSE and a slight increase in SSIM over the EINGN-NET model, suggesting the effectiveness of our algorithm. 
Moreover, based on the MSE, SSIM and DR for judging quality, it is shown that the HQSNet is enhanced by AA.
\begin{table}[htp]
\renewcommand{\arraystretch}{1.2}
\centering
\begin{tabular}{cccccc}
    \toprule
    & AA-HQSNet & HQSNet & EITGN-NET \cite{colibazzi2022learning} & GN-LM \cite{LIU2018304} & D-bar \cite{doi:10.1137/1.9781611972344} \\
    \hline
    MSE & \textbf{1.69}$\boldsymbol{\times10^{-3}}$ & 1.96$\times 10^{-3}$ & 2.04$\times 10^{-3}$ & 5.03$\times 10^{-3}$ & 6.85$\times 10^{-3}$ \\
    SSIM & \textbf{0.97} & 0.96 & 0.95 & 0.84 & 0.82 \\
    DR & 103 & 109 & 111 & \textbf{101} & 108 \\
    \bottomrule
\end{tabular}
\caption{Performance evaluation of AA-HQSNet, HQSNet, EITGN-Net, GN-LM, and D-bar in terms of MSE, SSIM, and DR Metrics.
}
\label{tab:per_comp}
\end{table}
 
\begin{figure}[htp]
\centering
\begin{minipage}[b]{0.83\linewidth}
    \subfigure[GT]{
        \begin{minipage}[b]{0.17\linewidth}
            \centering
            \includegraphics[width=.8\linewidth]{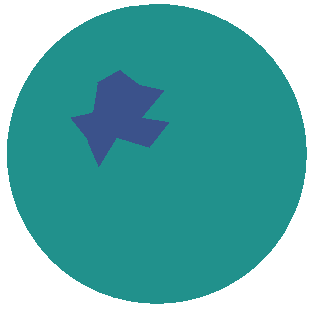}\vspace{4pt}
            \includegraphics[width=.8\linewidth]{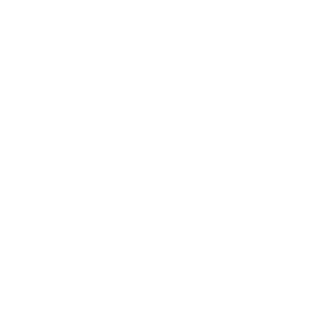}\vspace{0pt}
            \begin{center}
                \footnotesize EIEI:
            \end{center}
            \includegraphics[width=.8\linewidth]{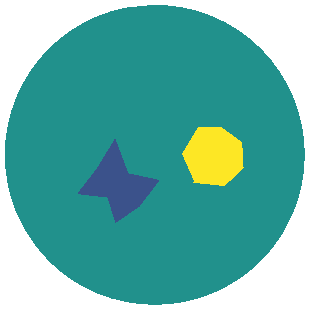}\vspace{2pt}
            \includegraphics[width=.8\linewidth]{figs/5.3_robustness_to_noise_noise54dB_block.png}\vspace{0pt}
            \begin{center}
                \footnotesize EIEI:
            \end{center}
            \includegraphics[width=.8\linewidth]{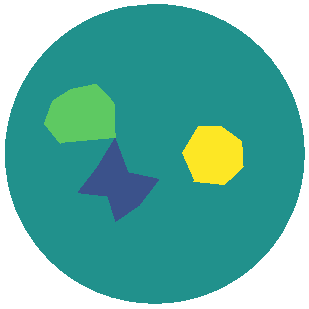}\vspace{2pt}
            \includegraphics[width=.8\linewidth]{figs/5.3_robustness_to_noise_noise54dB_block.png}\vspace{0pt}
            \begin{center}
                \footnotesize EIEI:
            \end{center}
            \includegraphics[width=.8\linewidth]{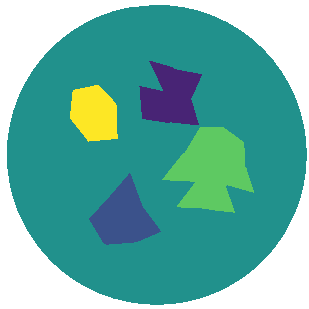}\vspace{2pt}
            \includegraphics[width=.8\linewidth]{figs/5.3_robustness_to_noise_noise54dB_block.png}\vspace{0pt}
            \begin{center}
               \footnotesize EIEI:
            \end{center}
        \end{minipage}
    }
    \hfill
    \subfigure[AA-HQSNet]{
        \begin{minipage}[b]{0.17\linewidth}
            \centering
            \includegraphics[width=.8\linewidth]{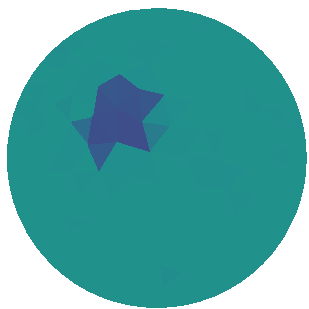}\vspace{2pt}
            \includegraphics[width=.8\linewidth]{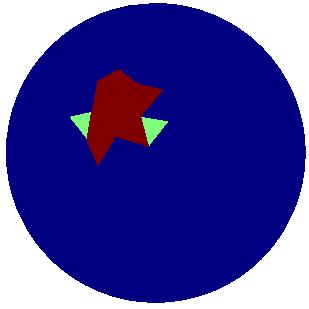}\vspace{0pt}
            \begin{center}
                \footnotesize 1.77
            \end{center}
            \includegraphics[width=.8\linewidth]{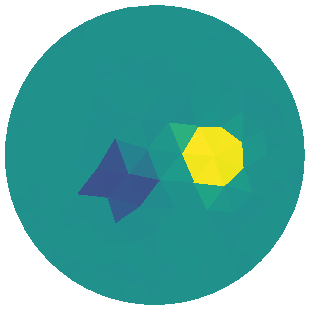}\vspace{2pt}
            \includegraphics[width=.8\linewidth]{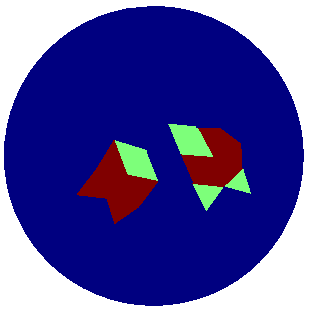}\vspace{0pt}
            \begin{center}
                \footnotesize 2.18
            \end{center}
            \includegraphics[width=.8\linewidth]{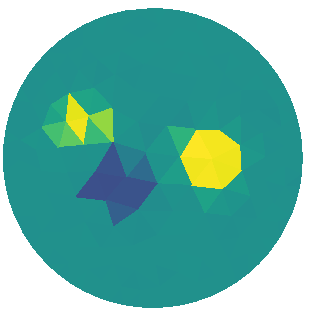}\vspace{2pt}
            \includegraphics[width=.8\linewidth]{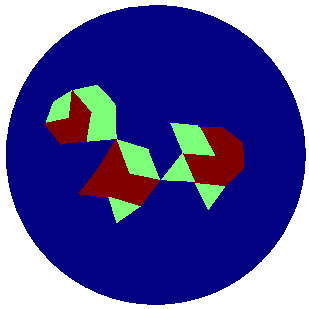}\vspace{0pt}
            \begin{center}
                \footnotesize 2.13
            \end{center}
            \includegraphics[width=.8\linewidth]{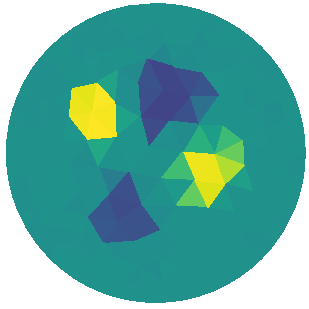}\vspace{2pt}
            \includegraphics[width=.8\linewidth]{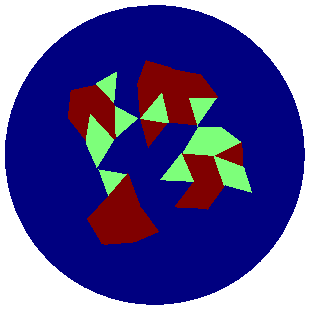}\vspace{0pt}
            \begin{center}
                \footnotesize 2.13
            \end{center}
        \end{minipage}
    }
    \hfill
    \subfigure[HQSNet]{
        \begin{minipage}[b]{0.17\linewidth}
            \centering
            \includegraphics[width=.8\linewidth]{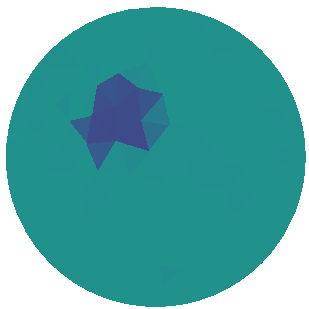}\vspace{2pt}
            \includegraphics[width=.8\linewidth]{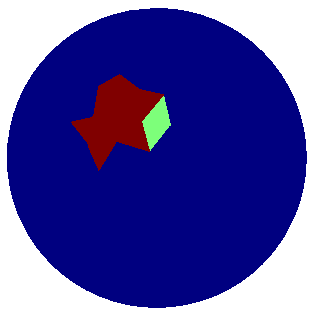}\vspace{0pt}
            \begin{center}
                \footnotesize 1.79
            \end{center}
            \includegraphics[width=.8\linewidth]{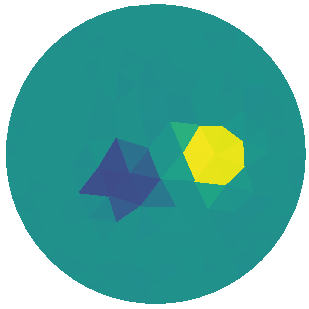}\vspace{2pt}
            \includegraphics[width=.8\linewidth]{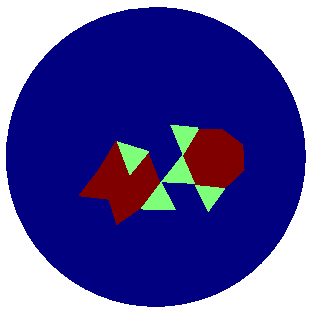}\vspace{0pt}
            \begin{center}
                \footnotesize 2.01
            \end{center}
            \includegraphics[width=.8\linewidth]{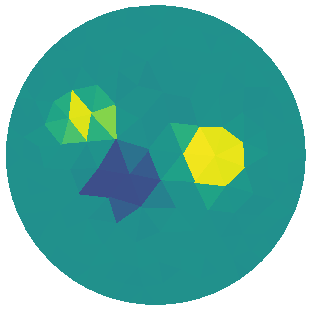}\vspace{2pt}
            \includegraphics[width=.8\linewidth]{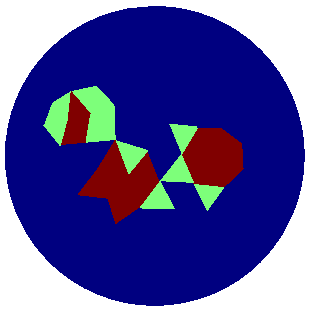}\vspace{0pt}
            \begin{center}
                \footnotesize 2.13
            \end{center}
            \includegraphics[width=.8\linewidth]{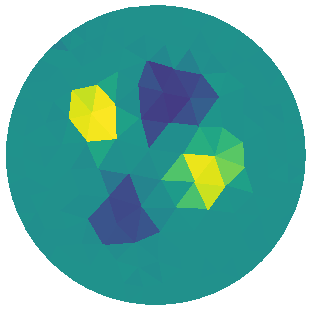}\vspace{2pt}
            \includegraphics[width=.8\linewidth]{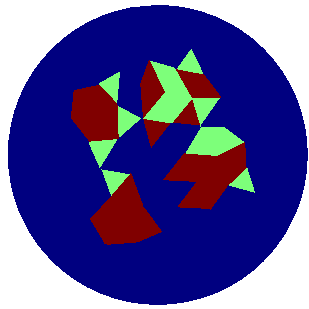}\vspace{0pt}
            \begin{center}
                \footnotesize 1.82
            \end{center}
        \end{minipage}
    }
    \hfill
    \subfigure[GN-LM]{
        \begin{minipage}[b]{0.17\linewidth}
            \centering
            \includegraphics[width=.8\linewidth]{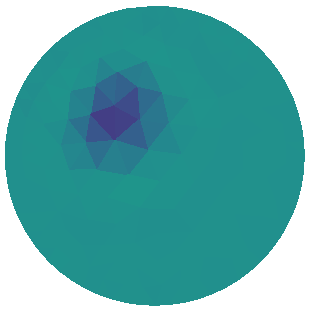}\vspace{2pt}
            \includegraphics[width=.8\linewidth]{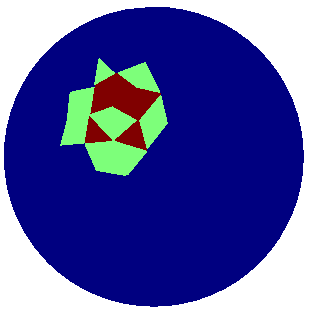}\vspace{0pt}
            \begin{center}
                \footnotesize 1.69
            \end{center}
            \includegraphics[width=.8\linewidth]{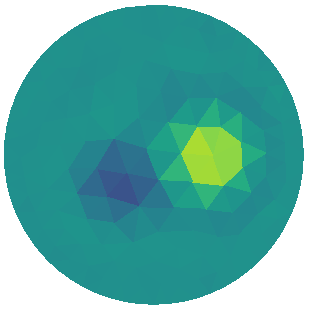}\vspace{2pt}
            \includegraphics[width=.8\linewidth]{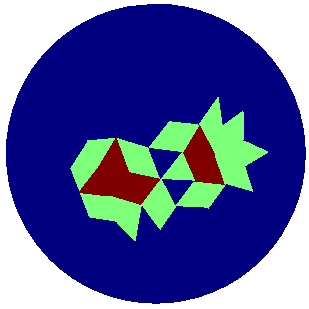}\vspace{0pt}
            \begin{center}
                \footnotesize 1.95
            \end{center}
            \includegraphics[width=.8\linewidth]{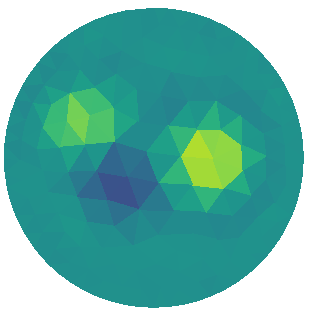}\vspace{2pt}
            \includegraphics[width=.8\linewidth]{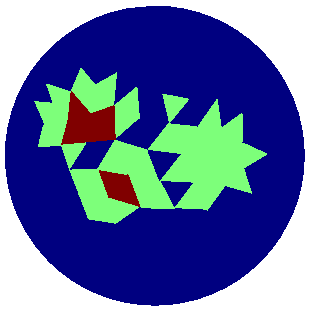}\vspace{0pt}
            \begin{center}
                \footnotesize 1.88
            \end{center}
            \includegraphics[width=.8\linewidth]{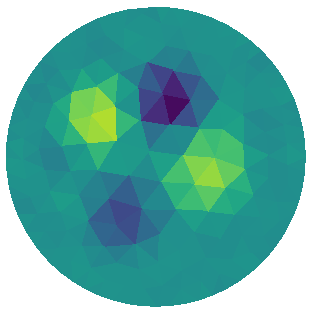}\vspace{2pt}
            \includegraphics[width=.8\linewidth]{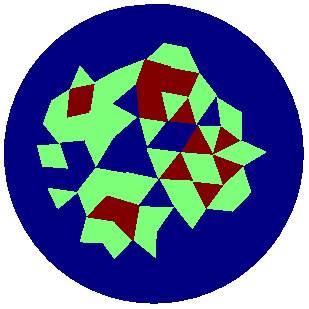}\vspace{0pt}
            \begin{center}
                \footnotesize 1.73
            \end{center}
        \end{minipage}
    }
    \hfill
    \subfigure[D-bar]{
        \begin{minipage}[b]{0.17\linewidth}
            \centering
            \includegraphics[width=.8\linewidth]{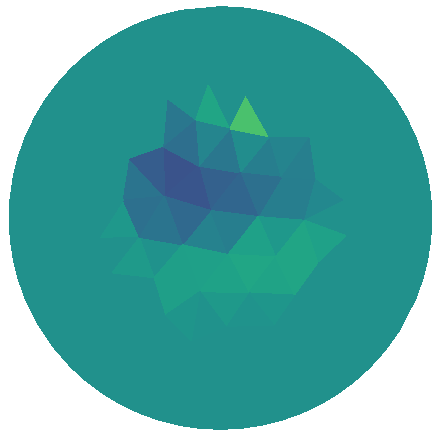}\vspace{2pt}
            \includegraphics[width=.8\linewidth]{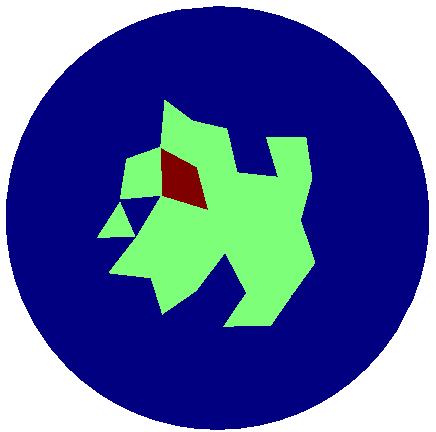}\vspace{0pt}
            \begin{center}
                \footnotesize 1.63
            \end{center}
            \includegraphics[width=.8\linewidth]{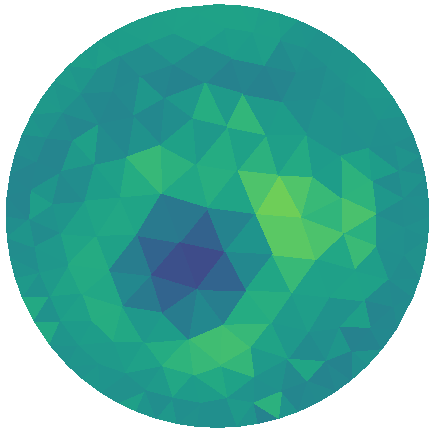}\vspace{2pt}
            \includegraphics[width=.8\linewidth]{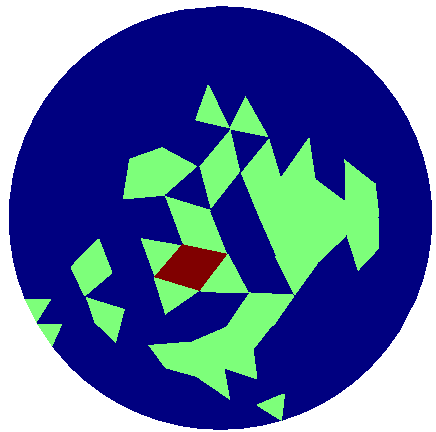}\vspace{0pt}
            \begin{center}
                \footnotesize 1.87
            \end{center}
            \includegraphics[width=.8\linewidth]{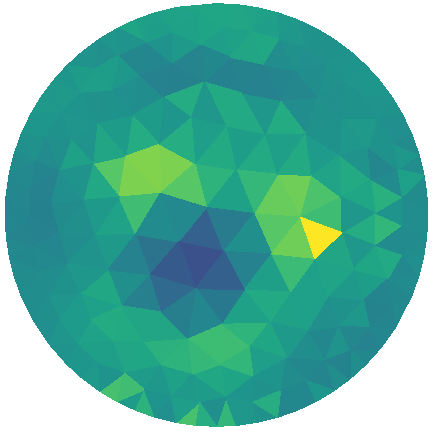}\vspace{2pt}
            \includegraphics[width=.8\linewidth]{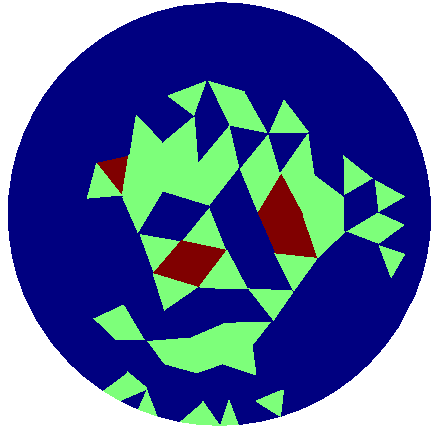}\vspace{0pt}
            \begin{center}
                \footnotesize 1.79
            \end{center}
            \includegraphics[width=.8\linewidth]{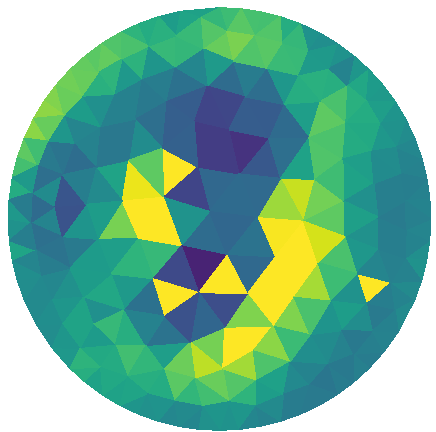}\vspace{2pt}
            \includegraphics[width=.8\linewidth]{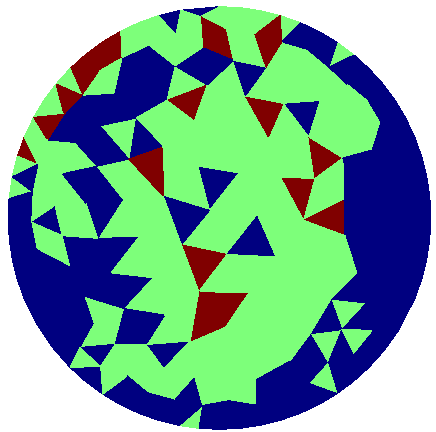}\vspace{0pt}
            \begin{center}
                \footnotesize 1.30
            \end{center}
        \end{minipage}
    }
\end{minipage}
\vfill
\caption{Visual reconstruction performance comparisons between AA-HQSNet, HQSNet, GN-LM, and D-bar with varying numbers of anomalies are shown in the top row. The associated structure maps (EIEI metric) are reported in the bottom row. The EIEI structure maps are color-coded, with yellow representing artifacts, red representing anomalies, and blue representing the background. The EIEI values for all test cases with varying numbers of anomalies ranging from 1 to 4 are reported, and a higher value of it indicates better performance}
\label{fig:comp1}
\end{figure}

The results are illustrated in Fig.\ref{fig:comp1}.
It is evident that AA-HQSNet generates sharper structures, fewer artifacts, and overall enhances the differentiation between anomalies.
As the number of anomalies increases, the performance of AA-HQSNet remains consistently favorable, whereas GN-LM exhibits a gradual decline in performance.

The positive influence of Anderson acceleration is noticeable to some extent, especially in cases with two anomalies present. Overall, AA-HQSNet effectively detects all anomalies with minimal artifacts and preserves their shape, as demonstrated by the EIEI values shown below each structure map in Fig. \ref{fig:comp1}.

\section{Algorithm investigation}\label{sec:algo-analysis}
In the previous section, we showcase the applications of our algorithm in solving the EIT inverse problem. To further validate the efficacy of our AA-HQSNet algorithm, we conduct additional experiments and analyses, presented in this section. Specifically, we focus on four key aspects of the algorithm:
\begin{enumerate*}[label=(\roman*)]
\item The selection of optimal hyperparameters such as $K^{(1)}$, $K^{(2)}$, $m^{(1)}$, $m^{(2)}$, $K$, and $\mu$;
\item Proximal neural networks and sensitivity analysis;
\item An ablation study to analyze the contribution of our AA-HQSNet algorithm;
\item Robustness analysis of the algorithm with varying levels of noise.
\end{enumerate*}

\subsection{Hyperparameters Analysis}
To investigate the impact of various hyperparameters, including $K^{(1)}$, $K^{(2)}$, $m^{(1)}$, $m^{(2)}$, $K$, and $\mu$, on the performance of the AA-HQSNet algorithm, we carry out a series of experimental investigations. Here, $K^{(1)}$ and $K^{(2)}$ represent the number of iterations in the Gauss-Newton-step and proximal-descent-step, respectively. $K$ denotes the number of folds of the unrolled AA-HQSNet, $m^{(1)}$ and $m^{(2)}$ refer to the depth of the AA algorithm (i.e., past periods that account for present information), and $\mu$ sets the degree of penalty defined in equation (\ref{7}).\par
To analyze the effects of $K^{(1)}$, $K^{(2)}$, and $K$ on the network, we divide the experiments into two groups. In the first group, we vary $K^{(1)}$ and $K^{(2)}$ within a range of 1 to 3 while keeping $K=8$ fixed. Note that although there are many combinations between $K^{(1)}$ and $K^{(1)}$, we simplify the combination by setting $K^{(1)}=K^{(2)}$. In contrast, in the second group, we keep $K^{(1)}$ and $K^{(2)}$ constant and manipulate $K\in \{4, 8, 12\}$.\par
Table \ref{tab:1} displays the results of the learned algorithm trained with different hyperparameter settings $(K^{(1)}, K^{(2)}, K)$ for simulated EIT. 
\begin{table}[htp]
\renewcommand{\arraystretch}{1.2}
\centering
\begin{tabular}{c|cccc}
    \toprule
    ($K$, $K^{(1)}$, $K^{(2)}$)  & MSE & PSNR & SSIM & DR \\
    \hline
    (8, 1, 1)\scriptsize(no AA) & 2.07$\times 10^{-3}$ & 32.77 & 0.935 & 108 \\
    (8, 2, 2) & \textbf{1.69} $\boldsymbol{\times10^{-3}}$ & \textbf{36.38} & \textbf{0.969} & \textbf{103} \\
    (8, 3, 3) & 1.72$\times 10^{-3}$ & 36.06 & 0.965 & 107 \\
    \hline
    (4, 2, 2) & 1.87$\times 10^{-3}$ & 34.41 & 0.956 & 114 \\
    (12, 2, 2) & 1.75$\times 10^{-3}$ & 35.78 & 0.965 & 103 \\
    \bottomrule
\end{tabular}
\caption{Numerical comparison of model performance with different hyperparameters values on ($K$, $K^{(1)}$, $K^{(2)}$). The results are obtained using datasets introduced in subsection \ref{ID}. All the experiments above are accelerated by AA with depth $m^{(1)}=m^{(2)}=2$ under a fixed number of iterations 30 except for the first row with hyperparameter groups (8,1,1)}
\label{tab:1}
\end{table}
 The performance gradually improves with increasing $K$ when $(K^{(1)}, K^{(2)})=(2,2)$ and tendes to reach peak performance when $K$ approaches 8. Similarly, the performance increases when $K^{(1)}$ and $K^{(2)}$ vary from 1 to 2 and decline as $K^{(1)}$ and $K^{(2)}$ continually increase to 3.
Therefore, The configuration $(K, K^{(1)}, K^{(2)})=(8,2,2)$ is a desirable choice as it strikes a balance between computational cost and resulting quality.\par
\begin{figure}[htp]
\begin{minipage}[b]{.6\linewidth}
    \centering
    \subfigure[][Variation in MSE under different $\mu$]{\includegraphics[scale=0.27]{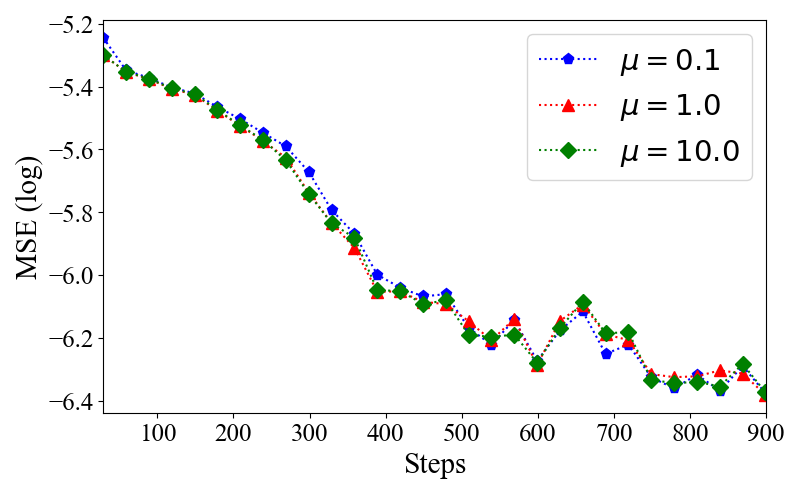}}
\end{minipage}
\medskip
 \begin{minipage}[b]{.4\linewidth}
    \centering
    \subfigure[][GT]{\includegraphics[scale=0.195]{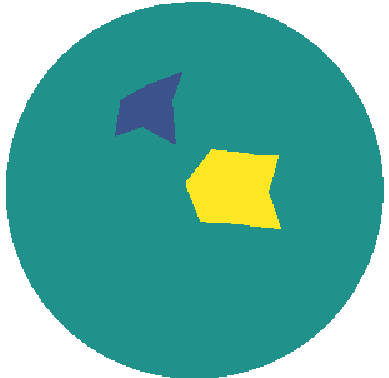}}
    \ \ \
    \subfigure[][$\mu = 0.1$]{\includegraphics[scale=0.245]{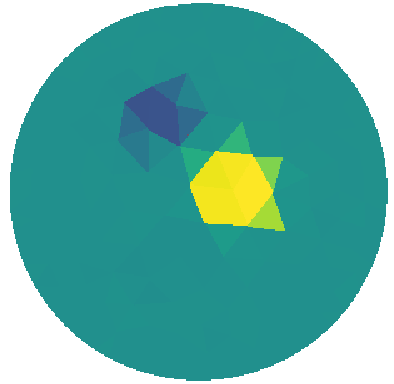}}
    
    \subfigure[][$\mu = 1.0$]{\includegraphics[scale=0.245]{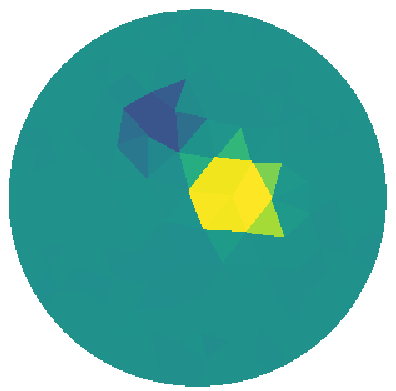}} \ \ \
    \subfigure[][$\mu = 10.0$]{\includegraphics[scale=0.245]{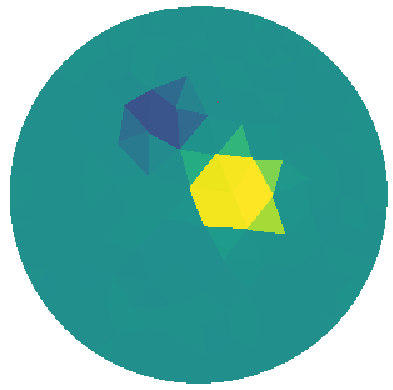}}
\end{minipage} 
\caption{Variation of $\mu$ under hyperparameter setting $(K, K^{(1)}, K^{(2)})=(8,2,2)$ with fixed number of iterations 30 and (b) ground truth of test sample; (c) reconstructed image with $\mu = 0.1$; (d) reconstructed image with $\mu=1.0$ and (e) reconstructed image with $\mu=10.0$}
\label{fig:mu}
\end{figure}

Fig.\ref{fig:mu} shows the quantitative variation of the mean squared error (MSE) for different values of $\mu$ under the same configuration $(K, K^{(1)}, K^{(2)})=(8, 2, 2)$.
 Our findings indicate that experimental performance does not significantly change as $\mu$ varies. This can be attributed to the incorporation of a neural network (NN) into our HQSNet algorithm, which stabilizes the influence of the penalty parameter $\mu$ by updating numerous parameters within the NN layers.

\subsection{The Sensitivity of the Proximal Neural Networks}
We investigate the impact of proximal neural network structures on the AA-HQSNet model, specifically, analyzing the effects of varying the hidden layers $L$ and the hidden size $N$ of the proximal neural networks $\Phi_{\theta}$. 
Here, we limit the ranges of the hidden layers to $L \in \{3,4,5\}$ and the hidden size to $N \in \{16, 32, 64\}$. Table~\ref{tab:2} presents the MSE values of AA-HQSNet trained with different network structures.  
Our findings indicate that, for a fixed $L$, the model performance remains similar, but for a fixed $N$,  the performance of the model varies. Specifically, when $N=64$, the MSE consistently declines as $L$ increases. Conversely, when $N=16$ or $32$, the MSE initially decreases and then increases as $L$ increases. Therefore, our results imply that the effect of $L$ on the model performance is more significant than that of $N$.
We have also examined the Structural Similarity Index (SSIM) and Dynamic Range (DR) values and find that the results are qualitatively similar to those presented in Table~\ref{tab:2}, so we omit those results.
\begin{table}[htp]
\renewcommand{\arraystretch}{1.2}
\setlength{\tabcolsep}{9pt}
\centering
\begin{tabular}{l|ccc}
\toprule
\diagbox{$L$}{MSE}{$N$} & 16 & 32 & 64 \\
\hline
3 & 2.57$\times 10^{-3}$ & 2.15$\times 10^{-3}$ & 2.06$\times 10^{-3}$ \\
\hline
4 & 1.77$\times 10^{-3}$ & {1.69} ${\times10^{-3}}$ & 1.77$\times 10^{-3}$ \\
\hline
5 & 1.84$\times 10^{-3}$ & 1.73$\times 10^{-3}$ & 1.70$\times 10^{-3}$ \\
\bottomrule
\end{tabular}
\caption{The effect of different combinations of layers $L$ and feature maps $N$ in the proximal gradient network under a fixed number of iterations 30}
\label{tab:2}
\end{table}

\subsection{Ablation Study}\label{ablation}
\begin{figure}[htp]
\centering
\includegraphics[width=11cm]{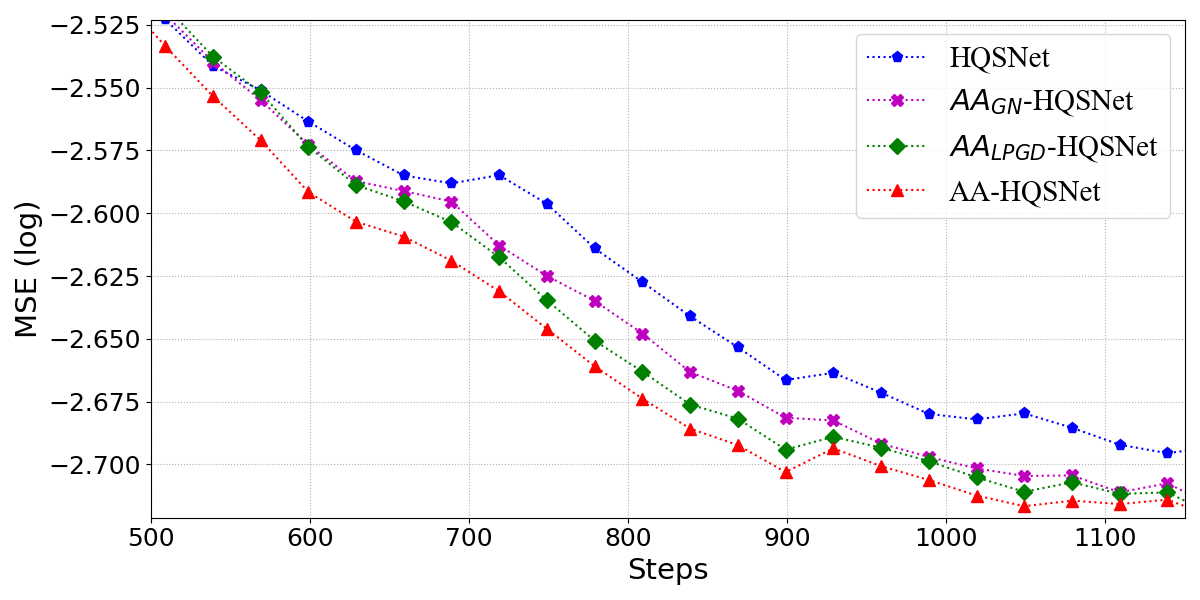}
\caption{The ablation study on AA-HQSNet with EIT within 40 iterations, and a smoothing factor of 0.85 is used to smooth out the curves for better visualization}
\label{fig:5.3}
\end{figure}

In this section, we present a detailed analysis of the effects of the AA component in our AA-HQSNet. To achieve this, we conduct ablation studies to systematically evaluate the performance of different variants of our proposed method. 
Specifically, we first introduce the base model, HQSNet, and then investigate the impact of applying Anderson acceleration to the Gauss-Newton (GN) step, which we refer to as $AA_{GN}$-HQSNet. We also experiment with the application of AA to the learned proximal gradient descent (LPGD) algorithm with the Gauss-Newton step without acceleration, which we call $AA_{LPGD}$-HQSNet. Finally, we evaluate our proposed method, as outlined in Algorithm \ref{alg:hqsaa}, which involves  applying the AA to both Gauss-Newton (GN) step and the learned proximal gradient descent (LPGD) step.
The evaluation results of the ablation models are summarized in Table \ref{table:2}. We observe significant differences in model performance when we separately accelerate the GN and LPGD steps. Specifically, we achieve a $3.1\%$ reduction in MSE when we individually accelerate the GN step compared to HQSNet. In contrast, we obtain a $10.2\%$ reduction in MSE when we individually accelerate the LPGD step. This difference is also evident for PSNR, as shown in Table \ref{table:2}. However, we do not observe any notable variation in SSIM and DR. When we apply both acceleration techniques, we achieve a $13.8\%$ reduction in MSE, which is much better than employing either acceleration technique individually. This improvement is also evident for other metrics. Overall, our ablation studies demonstrate that the AA component plays a crucial role in improving the performance of our AA-HQSNet.
\begin{table}[htp]
\renewcommand{\arraystretch}{1.2}
\centering
\begin{tabular}{c c|c|c|cccc}
    \toprule
    & Model & GN+AA & LPGD+AA & MSE & PSNR & SSIM & DR \\
    \hline
    \footnotesize &\ \ \ \ \ \ \ \ \ \ \ \ \ \ HQSNet & \red{\ding{56}} & \red{\ding{56}} & 1.96$\times 10^{-3}$ & 34.67 & 0.961 & 109 \\
    \footnotesize & \ \ \ \ $AA_{GN}$-HQSNet & \color{green}{\ding{52}} & \red{\ding{56}} & 1.90$\times 10^{-3}$ & 34.75 & 0.960 & 108 \\
    \footnotesize & \ $AA_{LPGD}$-HQSNet & \red{\ding{56}} & \color{green}{\ding{52}} & 1.76$\times 10^{-3}$ & 35.57 & 0.962 & 109 \\
    \footnotesize & \ \ \ \ \ \ \ \  AA-HQSNet & \color{green}{\ding{52}} & \color{green}{\ding{52}} & \textbf{1.69}$\boldsymbol{\times10^{-3}}$ & \textbf{36.38} & \textbf{0.969} & \textbf{103} \\
    \bottomrule
\end{tabular}
\caption{The performance of the reconstructions for EIT in terms of MSE, PSNR, SSIM, and DR metrics by ablation study within 30 iterations, where \red{\ding{56}} indicates the algorithm without Anderson acceleration and {\color{green}{\ding{52}}} indicates the algorithm with Anderson acceleration}
\label{table:2}
\end{table}
Additionally, we present visual comparisons in Fig.\ref{fig:5.3}, which displays the variation of MSE over iterations for all four HQSNet methods. 
Table \ref{table:2} and Figure \ref{fig:5.3} demonstrate that the AA-HQSNet method outperforms the other variants in terms of MSE, PSNR, SSIM, and DR scores. The AA-HQSNet algorithm reduces the MSE measurement to 0.86 times compared to the HQSNet algorithm with almost no extra time cost. Additionally, it decreases by 6 in the DR measurement. In summary, the AA-HQSNet proposed in this work achieves outstanding reconstruction performance, mainly attributed to the proposed AA algorithm.

\subsection{Robustness to Noise}
\begin{figure}[hpt]
\centering
\begin{minipage}[b]{0.73\linewidth}
    \subfigure[GT]{
        \begin{minipage}[b]{0.2\linewidth}
            \centering
            \includegraphics[width=.8\linewidth]{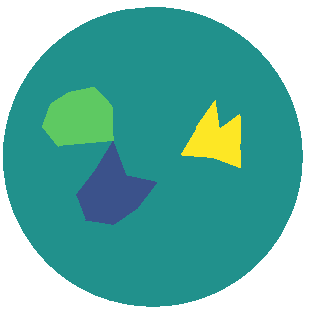}\vspace{4pt}
            \includegraphics[width=.8\linewidth]{figs/5.3_robustness_to_noise_noise54dB_block.png}\vspace{0pt}
            \begin{center}
                \footnotesize EIEI:
            \end{center}
            \vspace{12pt}
            \includegraphics[width=.8\linewidth]{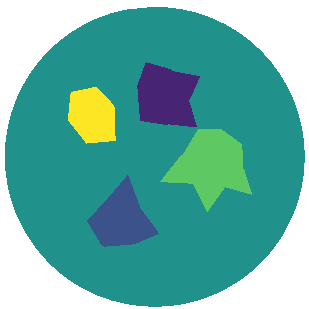}\vspace{2pt}
            \includegraphics[width=.8\linewidth]{figs/5.3_robustness_to_noise_noise54dB_block.png}\vspace{0pt}
            \begin{center}
                \footnotesize EIEI:
            \end{center}
            \vspace{12pt}
            \includegraphics[width=.8\linewidth]{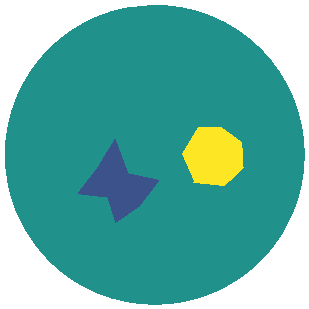}\vspace{2pt}
            \includegraphics[width=.8\linewidth]{figs/5.3_robustness_to_noise_noise54dB_block.png}\vspace{0pt}
            \begin{center}
                \footnotesize EIEI:
            \end{center}
        \end{minipage}
    }
     \hspace{0em}
    \subfigure[AA-HQSNet]{
        \begin{minipage}[b]{0.2\linewidth}
            \centering
            \includegraphics[width=.8\linewidth]{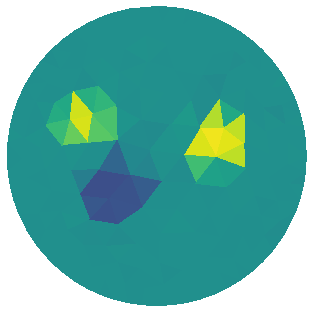}\vspace{2pt}
            \includegraphics[width=.8\linewidth]{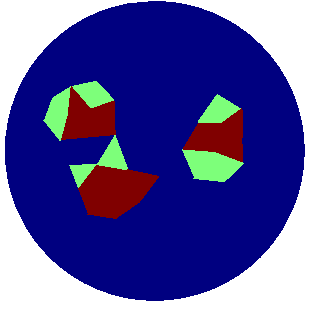}\vspace{0pt}
             \begin{center}
                \footnotesize 2.15
              \end{center}
            \vspace{12pt}
            \includegraphics[width=.8\linewidth]{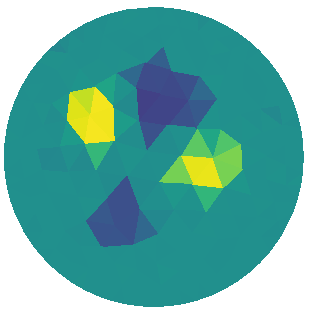}\vspace{2pt}
            \includegraphics[width=.8\linewidth]{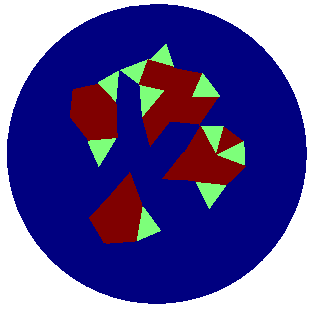}\vspace{0pt}
            \begin{center}
                \footnotesize 1.96
            \end{center}
            \vspace{12pt}
            \includegraphics[width=.8\linewidth]{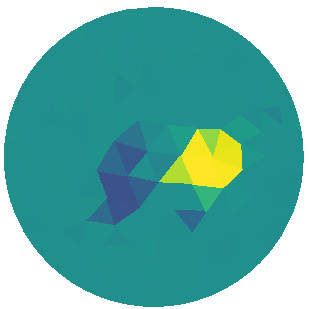}\vspace{2pt}
            \includegraphics[width=.8\linewidth]{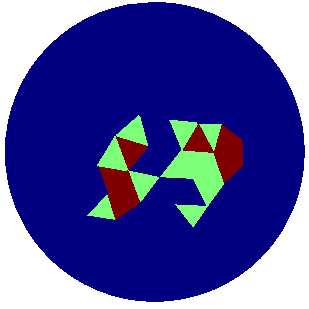}\vspace{0pt}
            \begin{center}
                \footnotesize 2.09
            \end{center}
        \end{minipage}
    }
    \hspace{0em}
    \subfigure[HQSNet]{
        \begin{minipage}[b]{0.2\linewidth}
            \centering
            \includegraphics[width=.8\linewidth]{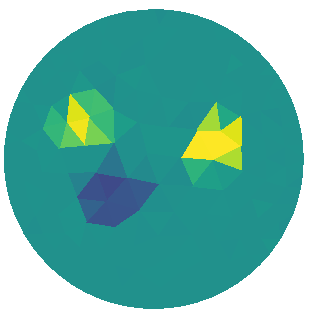}\vspace{2pt}
            \includegraphics[width=.8\linewidth]{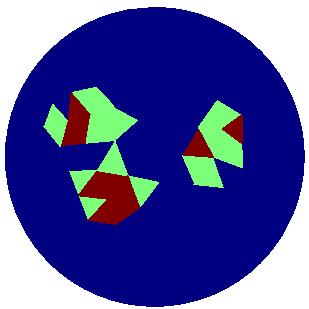}\vspace{0pt}
            \begin{center}
                \footnotesize 2.10
            \end{center}
            \vspace{12pt}
            \includegraphics[width=.8\linewidth]{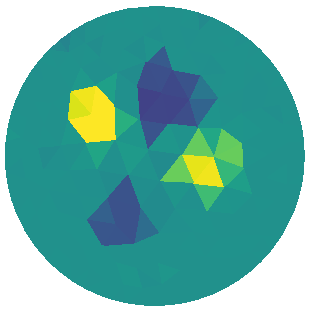}\vspace{2pt}
            \includegraphics[width=.8\linewidth]{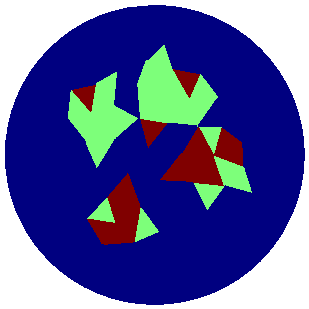}\vspace{0pt}
            \begin{center}
                \footnotesize 1.93
            \end{center}
            \vspace{12pt}
            \includegraphics[width=.8\linewidth]{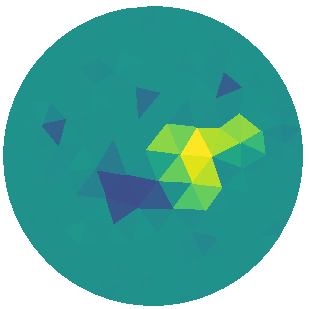}\vspace{2pt}
            \includegraphics[width=.8\linewidth]{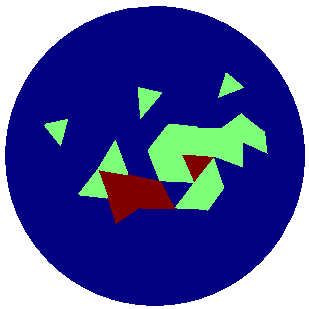}\vspace{0pt}
            \begin{center}
                \footnotesize 2.03
            \end{center}
        \end{minipage}
    }
    \hspace{0em}
    \subfigure[GN-LM]{
        \begin{minipage}[b]{0.2\linewidth}
            \centering
            \includegraphics[width=.8\linewidth]{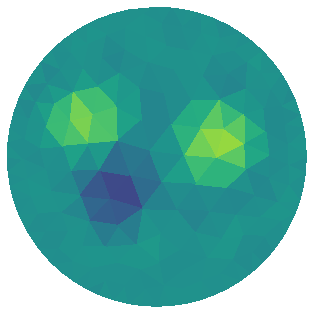}\vspace{2pt}
            \includegraphics[width=.8\linewidth]{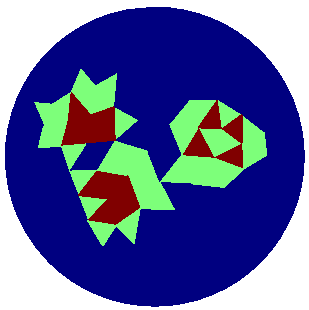}\vspace{0pt}
            \begin{center}
                \footnotesize 1.87
            \end{center}
            \vspace{12pt}
            \includegraphics[width=.8\linewidth]{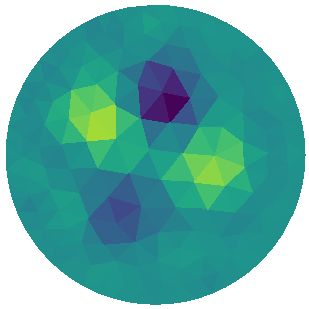}\vspace{2pt}
            \includegraphics[width=.8\linewidth]{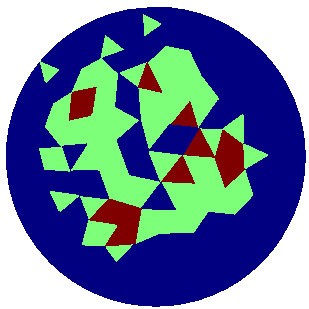}\vspace{0pt}
            \begin{center}
                \footnotesize 1.60
            \end{center}
            \vspace{12pt}
            \includegraphics[width=.8\linewidth]{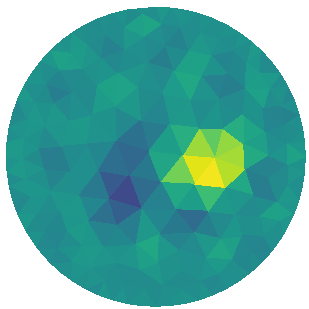}\vspace{2pt}
            \includegraphics[width=.8\linewidth]{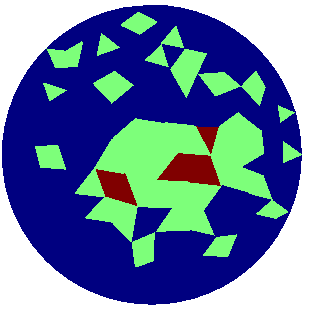}\vspace{0pt}
            \begin{center}
                \footnotesize 1.55
            \end{center}
        \end{minipage}
    }
\end{minipage}
\vfill
\caption{Visual comparison between $\sigma^{GT}$ and reconstructions obtained by AA-HQSNet (first column), HQSNet (second column), and GN-LM (third column) from measurements of different levels of noise. The reconstructed conductivities $\hat{\sigma}$ (first row) at a noise level of 54db and corresponding structure maps for EIEI metric (second row), at a noisier level of 48dB (third row) and of 42dB (fifth row) respectively, and we recall that for EIEI value, the higher, the better}
\label{fig:noise_eiei}
\end{figure}

We evaluate the performance of our proposed AA-HQSNet on the simulated EIT dataset comprising 50 degraded-clean voltage pairs corrupted by additive white Gaussian noise. Furthermore, we compare our method with several recent methods \cite{colibazzi2022learning, LIU2018304}.
To create the noisy measurements $v$, we add Gaussian noise specified by $\bar{n}\sim \mathcal{N}(0,s^{2})$, where the standard deviation $s$ is defined as $\eta\bar{v}$, with $\bar{v}$ being the average value. Correspondingly, the noisy model is represented by 
\begin{equation*}
v=v+ \eta \cdot \bar{v}\cdot rand \ (n_{M}).
\end{equation*}
We train the network with $\eta = 5\times 10^{-3}$, which corresponds to a Signal-to-Noise-Ratio (SNR) strength of 48dB, while testing with $\eta=2.5\times 10^{-3}$ (SNR=54dB) and more severe noise level $\eta=1\times 10^{-2}$ (SNR=42dB).
The evaluation results are reported in Table \ref{tab:4}.
\begin{table}[htp]
\renewcommand{\arraystretch}{1.2}
\centering
\begin{tabular}{ccccc}
\toprule
& AA-HQSNet & HQSNet & EITGN-NET \cite{colibazzi2022learning} & GN-LM \cite{LIU2018304} \\
\hline
noise 54dB & & & & \\
MSE & \textbf{2.1}$\boldsymbol{\times10^{-3}}$ & 2.3$\times 10^{-3}$ & 3.7$\times 10^{-3}$ & 5.5$\times 10^{-3}$ \\
SSIM & \textbf{0.96} & \textbf{0.96} & 0.90 & 0.75 \\
DR & \textbf{101} & 107 & 108 & 109 \\
\hdashline[1pt/1pt]
noise 48dB & & & & \\
MSE & \textbf{2.7}$\boldsymbol{\times10^{-3}}$ & 2.8$\times 10^{-3}$ & 4.3$\times 10^{-3}$ & 6.5$\times 10^{-3}$ \\
SSIM & \textbf{0.95} & \textbf{0.95} & 0.85 & 0.61 \\
DR & \textbf{105} & 110 & 114 & 117 \\
\hdashline[1pt/1pt]
noise 42dB & & & & \\
MSE & \textbf{6.4}$\boldsymbol{\times10^{-3}}$ & \textbf{6.4}$\boldsymbol{\times10^{-3}}$ & 7.0$\times 10^{-3}$ & 12$\times 10^{-3}$ \\
SSIM & \textbf{0.85} & \textbf{0.85} &  0.71 & 0.42 \\
DR & 121 & \textbf{119} &  126 & 131 \\
\bottomrule
\end{tabular}
\caption{The performance of the models on noisy measurements is evaluated based on the averaged MSE, SSIM, and DR metrics of the test samples with noise levels of 54dB, 48dB, and 42dB (more severe noise).
}
\label{tab:4}
\end{table}
Our method achieve significant performance in all three different noise level scenarios. In particular, our AA-HQSNet outperforms EITGN-NET with a $43.2\%$ reduction and $37.2\%$ reduction on noise levels 54dB and 48dB, respectively, in terms of MSE. These improvements are also evident for other metrics. The visual comparison is presented in Fig.\ref{fig:noise_eiei}, which includes three samples with three anomalies (first row), four anomalies (third row), and two anomalies (fifth row) individually.
From a visual comparison of the associated EIEI results, we can observe how the noise affects the quality of reconstruction. And note that a higher value of the EIEI indicator indicates better performance.
Clearly, the metrics reported in Table \ref{tab:4} confirm the significant overall performance of our proposed method in terms of robustness to noise compared to state-of-the-art methods.

\section{Discussion and conclusion}\label{sec:conclusion}
In this work, we proposed the learned half-quadratic splitting algorithm with Anderson acceleration (AA-HQSNet) to enhance the EIT image reconstruction. To develop our approach, we proposed the HQSNet algorithm that incorporates physical operators in the algorithm unrolling process. HQSNet consists of two subproblems: the Gauss-Newton and the learned proximal gradient descent steps. Embedding physics in deep neural networks can improve accuracy and robustness, thus improving safety in real-world applications. Subsequently, we employ the Anderson acceleration (AA) method for both subproblems, which can continue to reduce the occurrence of artifacts that commonly arise in ill-posed inverse problems.

A main perspective for future work is to scale the proposed methodology to handle very high-dimensional problems by extending our scheme to other iterative optimization algorithms such as the alternating direction method of multipliers (ADMM)~\cite{boyd2011distributed} and primal-dual~\cite{he2014convergence}.
Another approach is to substitute the auxiliary constraint condition $x=z$ with a data-driven condition $x=\mu(z)$ where $\mu(\cdot)$ is 
a deep generative model or a graphical neural network that learns from the training dataset. This approach can provide a sub-manifold constraint of the ambient space, leading to stronger regularization~\cite{herzberg2021graph,holden2022bayesian}; 
Additionally, it would be interesting to explore the theoretical properties of the resulting algorithm, including its convergence and upper and lower bounds~\cite{tang2022accelerating}.  
Future work could also study the application of the proposed methodology to other imaging problems, particularly those that are severely ill-posed and involve highly non-regular models such as in optical diffraction tomography.

\clearpage
\section*{Disclosures and Declaration}
We declare that we have no financial and personal relationships with other people or organizations that can inappropriately influence our work, there is no professional or other personal interest of any nature or kind in any product, service and/or company that could be construed as influencing the position presented in, or the review of, the manuscript entitled.

\section*{Data Availability Statement}
The datasets generated during and/or analysed during the current study are available in the ~\href{https://github.com/CCcodecod/AA-HQSNet}{https://github.com/CCcodecod/AA-HQSNet}.

\begin{acknowledgements}
The work is supported by the NSF of China (12101614) and the NSF of Hunan (2021JJ40715). 
We are grateful to the High Performance Computing Center of Central South University for assistance with the computations.
\end{acknowledgements}

\bibliographystyle{plain}      
\bibliography{a-reference}   

\end{document}